\documentclass[a4paper, onecolumn]{IEEEtran}

\PassOptionsToPackage{numbers, compress}{natbib}
\usepackage{natbib}
\usepackage[utf8]{inputenc} 
\usepackage[T1]{fontenc}    
\usepackage{hyperref}       
\usepackage{url}            
\usepackage{booktabs}       
\usepackage{amsfonts}       
\usepackage{nicefrac}       
\usepackage{microtype}      
\usepackage{color}
\usepackage[dvipsnames]{xcolor}       
\usepackage{amsmath, amssymb, amsthm}
\usepackage{mathtools}
\usepackage{bbm}
\usepackage{pgf}
\usepackage{wrapfig}

\newcommand{\PtildCal}{\tilde{\mathcal{P}}}
\newcommand{\Pcal}{\mathcal{P}}

\newcommand{\Xcal}{\mathcal{X}}
\newcommand{\Rbb}{\mathbb{R}}

\newcommand{\Nbb}{\mathbb{N}}
\newcommand{\Qbf}{\mathbf{Q}}
\newcommand{\Mcal}{\mathcal{M}}
\newcommand{\Rcal}{\mathcal{R}}
\newcommand{\Acal}{\mathcal{A}}

\newcommand{\Pcont}{\mathcal{C}}

\newcommand{\bracket}[1]{\left( #1 \right)}
\newcommand{\bracketSet}[1]{\left\{ #1 \right\}}

\newcommand{\fDiv}[2]{D_f \left( #1 \middle\Vert #2 \right)}
\newcommand{\fDivBin}[2]{D_f^{\mathrm{B}} \left( #1 \middle\Vert #2 \right)}
\newcommand{\clip}{\mathrm{clip}}

\newcommand{\klDiv}[2]{D_{\mathrm{KL}} \left( #1 \middle\Vert #2 \right)}
\newcommand{\tvDiv}[2]{D_{\mathrm{TV}} \left( #1 , #2 \right)}

\DeclareMathOperator*{\argmin}{arg\,min}

\theoremstyle{plain}
\newtheorem{theorem}{Theorem}[section]
\newtheorem{proposition}[theorem]{Proposition}
\newtheorem{lemma}[theorem]{Lemma}

\theoremstyle{definition}
\newtheorem{definition}[theorem]{Definition}

\theoremstyle{remark}
\newtheorem{remark}[theorem]{Remark}

\title{Exactly Minimax-Optimal Locally Differentially Private Sampling}

\author{
    Hyun-Young~Park$^\dagger$, Shahab~Asoodeh$^\ddagger$, and Si-Hyeon~Lee$^\dagger$\\
    $^\dagger$School of Electrical Engineering, KAIST\\
    $^\ddagger$Department of Computing and Software, McMaster University\\
    phy811@kaist.ac.kr; asoodeh@mcmaster.ca; sihyeon@kaist.ac.kr
}

\begin{document}

\maketitle

\begin{abstract}
  The sampling problem under local differential privacy has recently been studied with potential applications to generative models, but a fundamental analysis of its privacy-utility trade-off (PUT) remains incomplete. In this work, we define the fundamental PUT of private sampling in the minimax sense, using the $f$-divergence between original and sampling distributions as the utility measure. 
  We characterize the exact PUT for both finite and continuous data spaces under some mild conditions on the data distributions, and  propose sampling mechanisms that are universally optimal for all $f$-divergences. Our numerical experiments demonstrate the superiority of our mechanisms over baselines, in terms of theoretical utilities for finite data space and of empirical utilities for continuous data space. 
\end{abstract}

\section{Introduction}\label{sec:intro}

Privacy leakage is a pressing concern in the realm of machine learning (ML), spurring extensive research into privacy protection techniques \cite{Nasr23-ScaleExtractTrainDataFromProdLM, Wu24-UnveilSecPrivEthicChatGPT, Yan24-ProtDataPrivLLMSurvey, Li24-ShareToLeakDiffusionModel}. Among these, local differential privacy (LDP) \cite{Shiva_subsampling} stands out as a standard model  and has been deployed in industry, e.g., by Google \cite{erlingsson2014rappor,Prochlo}, Apple \cite{Apple_Privacy}, Microsoft \cite{ding2017collecting}.
In the LDP framework, individual clients
randomize their data on their own
devices and send it to a potentially untrusted aggregator for analysis, thus preventing the user data from being inferred. 
However, this perturbation inherently diminishes data utility.
Consequently, the central challenge in privacy mechanism design lies in optimizing utility while preserving the desired level of privacy protection. This goal involves characterizing the optimal balance between privacy parameter and utility, referred to as the privacy-utility trade-off (PUT). {The analysis of the PUT and the proposal of privacy mechanisms have been actively conducted for various settings of statistical inference and machine learning \cite{Duchi13-LDPStatMinmaxRate, Ye18-OptDiscreteDistEstLDP, Bhowmick19-ProtReconstAttactPrivFed, Asi22-PrivUnitOptim, Ghazi23-compPairwiseStatLDP, Lee23-MinmaxRiskOptEstFuncLDP, Asi23-FastOptLDPMeanEstandProj, Isik23-ExactOptPUCTDistMeanEst, Bassily14-PrivERM, Truex20-LDPFed, Ghazi23-RegressLDP, Feldman18-PrivAmpbyIter, Koyejo22-NetChangePointLocLDP, Chen22-DecentralWirelessFedDP, Allouah23-PrivRobUtilityTrilemma, Guo23-PrivAwareCompFed, Liu23-OnlineLDPQuantileInf, Sima24-onlineDistLearnLDP}.}

Most research in this field focuses on scenarios where each client has only a single data point. However, there are increasingly more applications where each client has a large local dataset with multiple data records. One can formulate the privacy requirement in these cases by assuming that clients have datasets of the same size, generated independently from an underlying distribution \citep{Girgis22-DistUserLevPrivMeanEst, Huang23-FedLinContBanditUserLevDP, Acharya23-userLevelLDPDiscreteEst, Bassily23-UserLevPrivStocConvOpt, Mao24-PrivShape, Sum24-PersonalizedPrivDistAI, Kent24-RateOptPhaseTransUserLevLDP}. 
This probabilistic assumption, however, restricts practical flexibility. 
The work \cite{Husain20-LDPSampling} explored a scenario where clients have large datasets that may vary in size and seek to privately release another dataset that closely resembles their original dataset.  In this scenario, local datasets can be represented by an empirical distribution, allowing each client to be seen as holding a probability distribution and generating a private sample from it.   
This setup, which is called \textit{private sampling}, is the main focus of this paper.

Private sampling has recently found applications in the private fine-tuning of large language models~\cite{Flemings24-DPNextTokPredLLM}. Additionally, private sampling is connected to the challenge of learning private generative models, a topic often explored in the central DP model \cite{Sampling_CentralDP_Smith,Ghazi23-DPSampleFromGaussAdnProdDist, Hamid16-SampPartDP, Berlingerio19-PrivSynDataDeepLearning, Liyang18-DPGAN, Xin20-PrivFLGAN, Torkzadehmahani19-DPCGAN, Liu19-PPGAN}. 
While there exist studies on private generative models within the local model \cite{Cunningham22-GeoPointGAN, Zhang23-LDPHighDimSynData, Shibata23-LDPImgGenFlowDeepGenModel, Gwon24-LDPGAN}, all these works assume a single data point per client. 
Very recently, the work \cite{Imola24-metricDPUserLev}  considered a setup where each client holds a probability distribution, but in a different context of query estimation.

The private sampling mechanism in \cite{Husain20-LDPSampling} can be described as follows. Initially, given a probability distribution $P$ representing a local dataset, the mechanism assumes a \textit{fixed} reference distribution. It then constructs what is termed the ``relative mollifier'',  a closed ball centered around the reference distribution with a radius equivalent to half of the privacy budget, within the space of probability distributions. 
Subsequently, the mechanism computes the projection of $P$ onto the relative mollifier, utilizing the Kullback-Leibler (KL) divergence. This projected distribution serves as the sampling distribution for generating a sample (see Section~\ref{subsec:relWorks} for more details). However, this mechanism has a notable shortcoming: the sampling distribution is only locally optimal within the relative mollifier. This, in turn, implies that the optimality of the sampling distribution depends on the choice of the reference distribution. 
A more fundamentally intriguing goal would be to formulate and characterize the PUT without such an ambiguity in the choice of reference distribution.

In this paper, we establish the optimality of locally private sampling in the minimax sense, and identify optimal private samplers. Our primary contributions are summarized as follows:
\begin{figure}[t]
    \centering
    \includegraphics[width=0.7\textwidth]{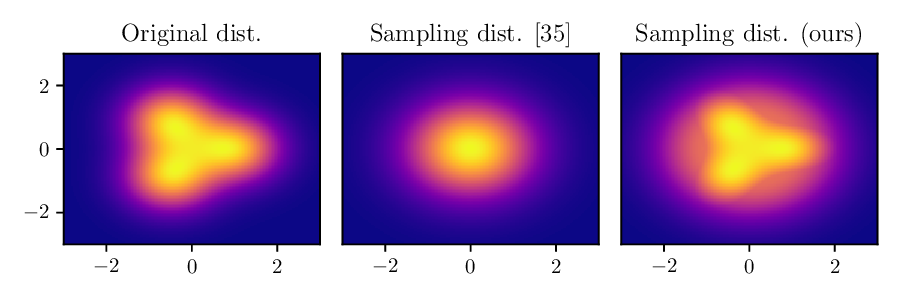}
    \caption{Original Gaussian ring distribution and the sampling distributions of the baseline  \cite{Husain20-LDPSampling} and our proposed mechanism for privacy budget $\epsilon=0.5$. The implementation details are in Appendix \ref{supp:numResultSetups}.
    } 
    
    \label{fig:ringGauss}
\end{figure}

\begin{itemize}
    \item The fundamental PUT of private sampling is rigorously defined in terms of minimax utility, which is commonly used in the literature of private estimation \cite{Duchi13-LDPStatMinmaxRate, Ye18-OptDiscreteDistEstLDP, Asi22-PrivUnitOptim, Acharya23-userLevelLDPDiscreteEst}. 
    We impose some minimal assumptions on client's distributions as in \cite{Ghazi23-DPSampleFromGaussAdnProdDist} (which studies sampling under \textit{central} DP, a weaker privacy model than the local model \cite{Dwork06-CalNoiseToSens}). 
    For utility measure, we use the \textit{$f$-divergence} \cite{Sason18-fDiv, SasonVerdu16-fDivIneqs} between the original and the sampling distributions, that includes KL divergence, total variation distance, squared Hellinger distance, and $\chi^2$-divergence as special cases.

    \item We characterize the exact PUT for both finite and continuous data spaces, and present optimal sampling mechanisms achieving the PUT.  Surprisingly, our mechanisms are \textit{universally optimal} under any choice of $f$-divergence for utility measure.    

    \item 
    We numerically demonstrate that our proposed mechanism outperforms the baseline method presented in \cite{Husain20-LDPSampling}. 
    Specifically, for finite data spaces, we derive a closed-form expression for the utility of both our mechanism and the baseline, allowing for an exact comparison of their utilities. 
    In the case of continuous data spaces, a closed-form expression for the baseline is not available, so we use empirical utility for the comparison. 
    Figure ~\ref{fig:ringGauss} illustrates our proposed mechanism outputs a distribution closer to the original, than the baseline.
\end{itemize}

All codes for experiments and figures are attached as a supplementary material{, and can be found at the online repository}\footnote{\url{https://github.com/phy811/Optimal-LDP-Sampling}.}.
The instructions to reproduce the results in the paper are in Appendix \ref{supp:reprodInst}.

\section{Problem Formulation}\label{sec:probForm}
\subsection{Notations and preliminaries}\label{subsec:notations}

\paragraph*{Notations}
For a sample space $\Xcal$, let $\Pcal(\Xcal)$ be the set of all probability distributions on $\Xcal$. 
For each $n \in \Nbb$, let $\Pcont(\Rbb^n)$ be the set of all continuous probability distributions on $\Rbb^n$.
For each positive integer $k \in \Nbb$, let $[k]:=\bracketSet{1,2,\cdots,k}$. For a subset $A \subset \Xcal$, $\mathbbm{1}_A:\Xcal \rightarrow \{0,1\}$ denotes the indicator function, defined as $\mathbbm{1}_A(x)=1$ for $x \in A$ and $\mathbbm{1}_A(x)=0$ for $x \notin A$. Also, for $s_1 \leq s_2$ and $x \in \mathbb{R}$, let $\clip(x; s_1, s_2) := \max\{s_1, \min\{s_2, x\} \}$. 
We refer to Appendix \ref{supp:measTheoryAssume} for the rigorous measure-theoretic assumptions underlying  the paper.
\paragraph*{$f$-divergence} For a convex function $f:(0,\infty) \rightarrow \mathbb{R}$ satisfying $f(1)=0$ and two probability distributions $P, Q \in \Pcal(\Xcal)$ on the same sample space $\Xcal$, let $\fDiv{P}{Q}$ denote the $f$-divergence \cite{Sason18-fDiv, SasonVerdu16-fDivIneqs}.
The general definition of $f$-divergence is given in Appendix \ref{supp:fDivReview}. 
For $\Xcal=\Rbb^n$ and $P,Q \in \Pcont(\Rbb^n)$ with $P \ll Q$ (that is, $P(A)=0$ whenever $Q(A)=0$), it is defined as
\begin{equation}
    \fDiv{P}{Q} = \int_{x:q(x)>0} q(x) {f\bracket{\frac{p(x)}{q(x)}}} dx, \label{eq:fdivDef}
\end{equation}

where $p,q$ are pdfs of $P,Q$, respectively, and we define $f(0)=\lim_{x \rightarrow 0^+}f(x) \in (-\infty,\infty]$. For finite $\Xcal$, we can replace the integral with the sum and replace $p,q$ with $P,Q$.
Several well-known distance measures between distributions are examples of $f$-divergence with different convex functions. 
For instance, KL divergence (relative entropy), total variation distance, squared Hellinger distance, and $\chi^2$-divergence are $f$-divergences with $f(x)=x\log x$, $f(x)=|x-1|/2$, $f(x)= (1-\sqrt{x})^2$, and $f(x) = x^2-1$, respectively. 
Two important properties of general $f$-divergence are keys for this work. First, $\fDiv{P}{Q} \geq 0$, and equality holds if $P=Q$. Furthermore, we have 
\begin{equation}
        \fDiv{P}{Q} \leq M_f := \lim_{x \rightarrow 0+} f(x) + xf(1/x), \label{eq:fDivMutSing}
    \end{equation} where equality holds if $P$ and $Q$ are mutually singular (that is, they have disjoint supports). 
For a more comprehensive list of such $f$-divergences and their properties, we refer the readers to \cite{SasonVerdu16-fDivIneqs}.

We denote the KL divergence and the total variation distance as $\klDiv{P}{Q}$ and $\tvDiv{P}{Q}$, respectively. 
We note that the total variation distance is in fact a metric on $\Pcal(\Xcal)$.

\subsection{System model}
Suppose a client has access to a distribution $P \in \Pcal(\Xcal)$ over a sample space $\mathcal{X}$, and wants to produce a sample in $\mathcal{X}$ which looks like being drawn from $P$ and to send it to a data curator.
We assume that there are some constraints on the possible data distribution $P$, so that $P$ is restricted to be in some subset $\PtildCal \subset \Pcal(\Xcal)$, and both the client and the curator know $\Xcal$ and $\PtildCal$. 
However, it is required that a sampled element does not leak the privacy about the original distribution $P$. 
For this purpose, the client and the curator agree a \textbf{private sampling mechanism} $\Qbf$, which is a conditional distribution from $\PtildCal$ to $\mathcal{X}$.
After that, the client produces a sample following the distribution $\Qbf(\cdot|P)$. 
To guarantee the privacy protection, we impose $\Qbf$ to satisfy the local differential privacy (LDP) \cite{Duchi13-LDPStatMinmaxRate, Husain20-LDPSampling}.
\begin{definition}
Let $\epsilon>0$.
A private sampling mechanism $\Qbf$ is said to satisfy \textbf{$\epsilon$-LDP}, or $\Qbf$ is an $\epsilon$-LDP mechanism, if for any $P,P' \in \PtildCal$ and $A \subset \Xcal$, we have
\begin{equation}
    \Qbf(A|P) \leq e^\epsilon \Qbf(A|P').
\end{equation}
\end{definition}
For convenience, for each $P \in \PtildCal$, let $\Qbf(P) \in \Pcal(\Xcal)$ denote the distribution of $X$ given $P$ through $\Qbf$, that is $\Qbf(P)(A)=\Qbf(A|P)$ for each $A \subset \Xcal$.
In this way, we equivalently see $\Qbf$ as a function $\Qbf:\PtildCal \rightarrow \Pcal(\Xcal)$. 
Let $\mathcal{Q}_{\Xcal,\PtildCal, \epsilon}$ denote the set of all $\epsilon$-LDP mechanisms $\Qbf:\PtildCal \rightarrow \Pcal(\Xcal)$.

As the utility loss of the private sampling, we use the $f$-divergence between the original distribution and the sampling distribution, $\fDiv{P}{\Qbf(P)}$.
Since the sampling procedure can be performed across many clients who may have different data distributions, we measure the utility loss of $\Qbf$ by the \textbf{worst-case $f$-divergence}, 
\begin{align}
    R_f(\Qbf) = \sup_{P \in \PtildCal} \fDiv{P}{\Qbf(P)}.
\end{align}
Given $\Xcal$, $\PtildCal$, $\epsilon$, and $f$, our goal is to find the smallest possible worst-case $f$-divergence,
\begin{align}
    \Rcal(\Xcal, \PtildCal, \epsilon, f) = \inf_{\mathbf{Q} \in \mathcal{Q}_{\Xcal,\PtildCal, \epsilon}} R_f(\mathbf{Q}),
\end{align}
and to find a mechanism $\mathbf{Q} \in \mathcal{Q}_{\Xcal,\PtildCal, \epsilon}$ achieving it.  We say that $\mathbf{Q} \in \mathcal{Q}_{\Xcal,\PtildCal, \epsilon}$ is \textbf{optimal} for $(\Xcal, \PtildCal, \epsilon)$ under $D_f$ if $R_f(\Qbf)=\Rcal(\Xcal, \PtildCal, \epsilon, f)$.

\subsection{Related work}\label{subsec:relWorks}
 
The most closely related work to our work is \cite{Husain20-LDPSampling}.
The system models are the same as this paper, except the formulation of PUT. 
They first \emph{fix} a reference probability distribution $Q_0 \in \Pcal(\Xcal)$, and only consider mechanisms $\Qbf$ satisfying $e^{-\epsilon/2} Q_0(A) \leq \Qbf(A|P) \leq e^{\epsilon/2} Q_0(A)$ for all $P \in \PtildCal$ and $A \subset \Xcal$. In other words, let $\Mcal_{\epsilon, Q_0} = \{Q \in \Pcal(\Xcal): e^{-\epsilon/2} Q_0(A) \leq Q(A) \leq e^{\epsilon/2} Q_0(A), \quad \forall A \subset \Xcal\}$. Then, they only consider $\Qbf$ such that $\Qbf(P) \in \Mcal_{\epsilon, Q_0}$. Note that this guarantees $\epsilon$-LDP. For each $P \in \PtildCal$, they sought to find $Q \in \Mcal_{\epsilon, Q_0}$ which minimizes $\klDiv{P}{Q}$, and set this $Q$ to be $\Qbf(P)$. First, they claimed to find a closed-form expression of such a minimizer $Q$ for finite $\Xcal$, given by
\begin{align}
    Q(x) = \clip \left(P(x)/r_P; e^{-\epsilon/2} Q_0(x), e^{\epsilon/2} Q_0(x)\right), \label{eq:prevMechDiscClosedFrom}
\end{align}
where $r_P > 0$ is a constant depending on $P$ ensuring $\sum_{x \in \Xcal} Q(x) = 1$.
Second, they presented an algorithm, called Mollified Boosted Density Estimation (MBDE), to approximate the optimal solution for continuous $\Xcal$. However, the utility varies over the choice of reference distribution $Q_0$, and they left the question of choosing a best $Q_0$ to achieve the best performance in both practice and theory. Moreover, we found that the closed-form in \eqref{eq:prevMechDiscClosedFrom} is incomplete, because for some $(P,Q_0)$, there may be no $r_P >0$ such that the RHS of \eqref{eq:prevMechDiscClosedFrom} does not sum to one. As an example, when $P,Q_0$ are point masses at different points, then we can easily see that the sum of \eqref{eq:prevMechDiscClosedFrom} is $e^{-\epsilon/2}$ for any $r_P>0$.

\section{Main Results}\label{sec:mainResult}

\subsection{Optimal private sampling over finite space}\label{subsec:optMechFinSpace}
First, we consider the finite case, where $\Xcal = [k]$ for some $k \in \Nbb$. A natural setup for $\PtildCal$ is that $\PtildCal = \Pcal([k])$, i.e. there is no restriction on the client distribution $P \in \Pcal([k])$.
In this case, we completely characterize the optimal worst-case $f$-divergence $\Rcal(\Xcal, \PtildCal, \epsilon, f)$ and find an optimal private sampling mechanism. Surprisingly, for each $k \in \Nbb$ and $\epsilon>0$, we found a single mechanism which is universally optimal for every $f$-divergence.
\begin{theorem}\label{thm:optMechFinite}
For each $k \in \Nbb$, $\epsilon>0$, and an $f$-divergence $D_f$, we have
\begin{equation}
    \mathcal{R}([k],\Pcal([k]), \epsilon, f) = \frac{e^\epsilon}{e^\epsilon+k-1}f\left(\frac{e^\epsilon+k-1}{e^\epsilon}\right) + \frac{k-1}{e^\epsilon+k-1}f(0). \label{eq:worstDeviationFinCase}
\end{equation}
Moreover, the mechanism $\Qbf^*_{k,\epsilon}$ constructed as below satisfies $\epsilon$-LDP and is optimal for $(\Xcal = [k], \PtildCal=\Pcal([k]), \epsilon)$ under any $D_f$:
\begin{equation}
    \Qbf^*_{k,\epsilon}(x|P) = \max\left(\frac{1}{r_P} P(x), \frac{1}{e^\epsilon+k-1}\right) \quad \forall x \in [k], P \in \Pcal([k]), \label{eq:optMechFormFin}
\end{equation}
where $r_P > 0$ is a constant depending on $P$ so that $\sum_{x=1}^{k} \Qbf^*_{k,\epsilon}(x|P)=1$. Furthermore, $r_P$ can be chosen such that $1 \leq r_P \leq (e^\epsilon+k-1)/e^\epsilon$.
\end{theorem}

{By definition, we have $\Qbf^*_{k,\epsilon}(x|P) \geq \frac{1}{e^\epsilon+k-1}$ for all $x \in \Xcal$. This also implies that $\Qbf^*_{k,\epsilon}(x|P) = 1 - \sum_{x' \in \Xcal \backslash \{x\}} \Qbf^*_{k,\epsilon}(x'|P) \leq 1-\frac{k-1}{e^\epsilon+k-1} = \frac{e^\epsilon}{e^\epsilon+k-1}$. Hence, $\frac{1}{e^\epsilon+k-1} \leq \Qbf^*_{k,\epsilon}(x|P) \leq \frac{e^\epsilon}{e^\epsilon+k-1}$. This clearly implies that $\Qbf^*_{k,\epsilon}$ satisfies $\epsilon$-LDP.}

\begin{figure}[htbp]
    \centering
        \includegraphics[width=0.5\textwidth]{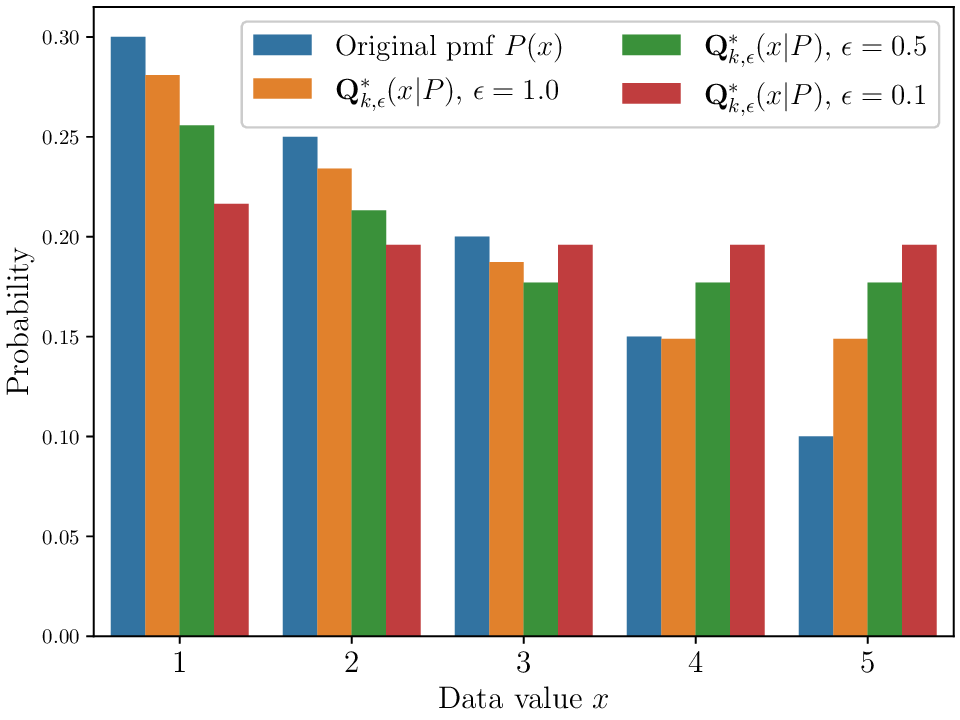}
        \caption{A visualization of the mechanism $\Qbf^*_{k,\epsilon}$}
        \label{fig:discMechVisual}
\end{figure}

\paragraph*{Behaviors of the optimal mechanism}
Let us observe some behaviors of the proposed mechanism with respect to the system parameters, whose formal proofs are in Appendix \ref{supp:mechBehavior}. 
We visualize how the mechanism $\Qbf^*_{k,\epsilon}$ works for different $\epsilon$ in Figure \ref{fig:discMechVisual}. {Here, we write $\mathcal{R}$ to mean $\mathcal{R}([k],\Pcal([k]), \epsilon, f)$ for simplicity.}
If $f(0)=\infty$, then $\mathcal{R}=\infty$, which means that $R_f(\Qbf)=\infty$ for any $\epsilon$-LDP sampling mechanism $\Qbf$. 
(Such a phenomenon happens for general $(\Xcal, \PtildCal)$, whenever $\PtildCal$ contains two mutually singular distributions) 
Hence, from now on in this paragraph, we assume $f(0)<\infty$. 
We can observe that $\mathcal{R}$ is decreasing in $\epsilon$ and increasing in $k$. 
For a fixed $k$, we have $\mathcal{R} \rightarrow 0$ as $\epsilon \rightarrow \infty$, which makes sense since $\epsilon \rightarrow \infty$ corresponds to the non-private case. 
Also, as $\epsilon \rightarrow 0$, we have $\Qbf^*_{k,\epsilon}(x|P) \rightarrow 1/k$ for every $P \in \Pcal([k])$ and $x \in [k]$, that is, $\Qbf^*_{k,\epsilon}(P)$ tends to the uniform distribution over $[k]$ for every $P \in \Pcal([k])$.
This fact can be also observed by Figure \ref{fig:discMechVisual}.

\paragraph*{Remarks about the constant $r_P$}
The value of $r_P$ may not be unique, but the mechanism $\Qbf^*_{k,\epsilon}$ does not depend on the choice of $r_P$. To see this, let us fix $P$, and let $g_r(x)=\max\left(\frac{1}{r} P(x), \frac{1}{e^\epsilon+k-1}\right)$. Suppose that $\sum_{x=1}^{k} g_{r}(x) = \sum_{x=1}^{k} g_{r'}(x) = 1$ for $r \leq r'$. Since $g_{r}(x) \geq g_{r'}(x)$ for each $x \in [k]$, the equality $\sum_{x=1}^{k} g_{r}(x) = \sum_{x=1}^{k} g_{r'}(x)$ implies that $g_{r}(x) = g_{r'}(x)$ for all $x \in [k]$. Hence $\Qbf^*_{k,\epsilon}$ is uniquely determined. 
Since $r \mapsto \sum_{x=1}^{k} g_{r}(x)$ is non-increasing and continuous, we can use the bisection method to find $r_P$. The `Furthermore' part of the theorem statement precisely means that for $r=1$ and $r=(e^\epsilon+k-1)/e^\epsilon$, the value of $\sum_{x=1}^{k} g_{r}(x)$ is at least and at most $1$, respectively, so that we can perform the bisection method with these two initial endpoints to find $r$ such that $\sum_{x=1}^{k} g_{r}(x)=1$.

\paragraph*{Comparison with the previous work \cite{Husain20-LDPSampling}}
The expression of the optimal mechanism in~\eqref{eq:optMechFormFin} is similar to \eqref{eq:prevMechDiscClosedFrom}, the expression of the KL divergence projection onto $\Mcal_{\epsilon, Q_0}$ derived by \citet{Husain20-LDPSampling}. Since $\frac{1}{e^\epsilon+k-1} \leq \Qbf^*_{k,\epsilon}(x|P) \leq \frac{e^\epsilon}{e^\epsilon+k-1}$, we can alternatively write $\Qbf^*_{k,\epsilon}(x|P) = \clip \left(\frac{1}{r_P} P(x); \frac{1}{e^\epsilon+k-1}, \frac{e^\epsilon}{e^\epsilon+k-1}\right)$. Hence, our optimal mechanism can be viewed as an instance of a generalized version of \eqref{eq:prevMechDiscClosedFrom},
where $Q_0$ is a positive measure, not necessarily a probability measure summing to one, given by $Q_0(x)=\frac{e^{\epsilon/2}}{e^\epsilon+k-1}$. 
{
A natural question is {whether $\Qbf_{k,\epsilon}^*(P)$ is a projection of $P$ onto $\Mcal_{\epsilon, Q_0}$, that is}
$\Qbf_{k,\epsilon}^*(P)$ is a minimizer of $\klDiv{P}{Q}$ among $Q \in \Mcal_{\epsilon, Q_0}$, where $\Mcal_{\epsilon, Q_0}$ is similarly defined as in Section~\ref{subsec:relWorks}.
As we shall discuss in Section \ref{sec:proofIdea}, this statement is true ---quite surprisingly--- even when we replace $D_{\mathrm{KL}}$ with any other $f$-divergences.
}
However,
our analysis {is more involved, as we need to show the optimality of the proposed mechanism over \emph{any other possible mechanisms}, including {minimizers with respect to other choices of $Q_0$}.}
Also, in Section \ref{sec:numResult}, we compare the worst-case $f$-divergence of our optimal mechanism with that of the mechanism proposed in \cite{Husain20-LDPSampling} which restricts $Q_0$ to be a probability distribution.

\subsection{Optimal private sampling over continuous space} \label{subsec:contSpace}

Next, we consider the continuous case, where $\Xcal = \Rbb^n$ for some $n \in \Nbb$. 
Some of the natural setups for $\PtildCal$ are (i) $\PtildCal = \Pcal(\Rbb^n)$, or (ii) $\PtildCal=\Pcont(\Rbb^n)$. We can also think some restrictive but still reasonable setups, such as the setups where (iii) $\PtildCal$ is the set of empirical distributions supported on some non-empty open subset of $\Rbb^n$, or where (iv) $\PtildCal$ is the set of continuous distributions on $[-1,1]^n$ having smooth pdf and zero mean.
However, we show that for a general class of $\PtildCal$ including these four cases, any $\epsilon$-LDP sampling mechanisms have the worst-case $f$-divergence equal to the maximum value $M_f$ of the $f$-divergence defined in \eqref{eq:fDivMutSing}. In the following proposition, $\Xcal$ may be a general sample space, not necessarily $\Rbb^n$ or finite space. The proof is in Appendix \ref{supp:remainPropStateAndProof}.
\begin{proposition}\label{prop:WorstFDivAlwaysMax}
Suppose that $\PtildCal$ contains infinitely many distributions which are pairwise mutually singular. Then, for any $\epsilon>0$, for any $\epsilon$-LDP mechanism $\mathbf{Q} \in \mathcal{Q}_{\Xcal,\PtildCal, \epsilon}$, and for any $f$-divergence, we have $R_{f}(\Qbf) = M_f$, where $M_f$ is define at \eqref{eq:fDivMutSing}. 
\end{proposition}

Hence, we need to consider sufficiently regular but practical setups for $\PtildCal$.

\paragraph*{Our setup}
 In this subsection, when $P,Q \in \Pcont(\Rbb^n)$, the corresponding small letters $p,q$ denote their pdfs.
In this paper, we consider the case that
\begin{equation}
    \PtildCal = \PtildCal_{c_1, c_2, h} := \{P \in \Pcont(\Rbb^n): c_1 h(x) \leq p(x) \leq c_2 h(x), \quad \forall x \in \Rbb^n\} \label{eq:possibleDensityAssumption}
\end{equation}
for some pre-known $h:\Xcal \rightarrow [0,\infty)$ such that $\int_{\Rbb^n} h(x)dx < \infty$ and $c_2 > c_1 \geq 0$.
Some of the sampling tasks and generative models in literature \cite{Goodfellow20-GAN, Husain20-LDPSampling} assume that the set of possible data distributions $\PtildCal$ satisfies $\PtildCal \subset \PtildCal_{c_1, c_2, h}$ for some $c_1, c_2, h$ satisfying the aforementioned condition, hence \eqref{eq:possibleDensityAssumption} is a moderate assumption. 
One example is the sampling from a Gaussian mixture, which is one of the canonical sampling tasks in literature \cite{Goodfellow20-GAN, Husain20-LDPSampling}.
Suppose that $\PtildCal$ consists of Gaussian mixtures, where each Gaussian has mean within a unit ball centered at the origin and has unit covariance. 
That is, $\PtildCal = \{\sum_{i=1}^{k} \lambda_i \mathcal{N}(\mu_i, I_{n}): k \in \Nbb, \lambda_i \geq 0, \sum_{i=1}^{k}\lambda_i = 1, \Vert \mu_i \Vert \leq 1\}$, where $I_n$ is the identity matrix of size $n \times n$. 
In this case, we can observe that $\PtildCal \subset \PtildCal_{0,1,h}$ for $h(x)=(2\pi)^{-n/2} \exp(-(\max(0, \Vert x \Vert - 1))^2/2)$, and it can be easily shown that $\int_{\mathbb{R}^n} h(x) < \infty$.

Without loss of generality, we may assume the following normalization condition on $c_1,c_2,h,\epsilon$:
\begin{gather}
   \int_{\Rbb^n} h(x)dx = 1, \quad c_1<1<c_2, \quad c_2 > c_1 e^\epsilon. \label{eq:normalizationCond}
\end{gather}
The reason is as follows. First, if any one of three inequalities $\int_{\Rbb^n} h(x)dx > 0$, $c_1 \int_{\Rbb^n} h(x)dx < 1$, and $c_2 \int_{\Rbb^n} h(x)dx > 1$ is not satisfied, then $\PtildCal_{c_1, c_2, h}$ is either an empty set or a singleton that consists of a distribution having pdf $c_1 h(x)$ or $c_2 h(x)$, which makes the problem trivial. Hence we impose all of the three inequalities. Then, we can normalize $c_1,c_2,h$, to make $\int_{\Rbb^n} h(x)dx = 1$ and $c_1<1<c_2$.
Furthermore, if $c_2 \leq e^\epsilon c_1$, then for any $P_1, P_2 \in \PtildCal_{c_1, c_2, h}$, we have $p_1(x)/p_2(x) \leq c_2/c_1 \leq e^\epsilon$, hence we can easily observe that the mechanism $\Qbf$ defined as $\Qbf(P)=P$ for all $P \in \PtildCal_{c_1, c_2, h}$ satisfies $\epsilon$-LDP and $R_f(\Qbf)=0$, hence the problem also becomes trivial, giving $\mathcal{R}(\Rbb^n,\PtildCal_{c_1, c_2, h}, \epsilon, f)=0$. Hence, we may assume \eqref{eq:normalizationCond}.

\paragraph*{Minimax utility and optimal mechanism} For the aforementioned setup, we can completely characterize $\mathcal{R}(\mathcal{X}, \PtildCal, \epsilon, f)$ and find a mechanism which is universally optimal for every $f$-divergence. The formula is similar to the discrete case, with a carefully chosen clipping bound.
\begin{theorem}\label{thm:optMechCont}
    For each $c_2>c_1 \geq 0, \epsilon>0$, and $h:\mathbb{R}^n \rightarrow [0,\infty)$ satisfying the normalization condition \eqref{eq:normalizationCond}, let us define the following constants determined by $c_1,c_2,\epsilon$:
    \begin{gather}
        b = \frac{c_2 - c_1}{(e^\epsilon-1)(1-c_1)+c_2-c_1},\quad
        r_1 = \frac{c_1}{b}, 
        \quad r_2 = \frac{c_2}{b e^\epsilon}.
    \end{gather}
    Then, we have
    \begin{align}
    \mathcal{R}(\Rbb^n,\PtildCal_{c_1, c_2, h}, \epsilon, f) &= \frac{1-r_1}{r_2-r_1} f(r_2) + \frac{r_2 - 1}{r_2 - r_1} f(r_1).
     \label{eq:worstDeviationBddDensity}
\end{align}
    Moreover, the mechanism $\Qbf^*_{c_1,c_2,h,\epsilon}$ constructed as below satisfies $\epsilon$-LDP and is optimal for $(\Xcal = \Rbb^n, \PtildCal=\PtildCal_{c_1, c_2, h}, \epsilon)$ under any $D_f$:

    For each $P \in \PtildCal$, $\Qbf^*_{c_1,c_2,h,\epsilon}(P)=:Q$ is defined as a continuous distribution with pdf
    \begin{equation}
        q(x) = \clip\left(\frac{1}{r_P} p(x); bh(x),  be^\epsilon h(x)\right), \label{eq:contMech}
    \end{equation}
    where $r_P>0$ is a constant depending on $P$ so that $\int_{\Rbb^n} q(x)dx = 1$. Furthermore, $r_P$ can be chosen such that $r_1 < r_P \leq r_2$.
\end{theorem}
It is also clear that $\Qbf^*_{c_1,c_2,h,\epsilon}$ satisfies $\epsilon$-LDP. Also, we note that $c_1<b<1<be^\epsilon<c_2$ and $0 \leq r_1 < 1 < r_2$, which is shown during the proof of Theorem \ref{thm:optMechCont} in Appendix \ref{supp:mainThmProof}. 

In practical scenario, it may be hard to expect $\PtildCal=\PtildCal_{c_1,c_2,h}$ exactly, and we may only know $\PtildCal \subset \PtildCal_{c_1,c_2,h}$ for some $c_1,c_2,h$ satisfying aforementioned conditions. In such case, we still propose to use $\Qbf^*_{c_1,c_2,h,\epsilon}$, and in Section \ref{sec:numResult}, we numerically show that this proposed mechanism is better than previously proposed mechanism \cite{Husain20-LDPSampling} in terms of the worst-case $f$-divergence.

\paragraph*{Behavior of the optimal mechanism}
We also observe some behaviors of the proposed mechanism with respect to the system parameters, whose formal proofs are in Appendix \ref{supp:mechBehavior}. 
{Again, we write $\mathcal{R}$ to mean $\mathcal{R}(\Rbb^n,\PtildCal_{c_1, c_2, h}, \epsilon, f)$ for simplicity.}
For a fixed $(c_1,c_2)$, $\mathcal{R}$ is decreasing in $\epsilon$. 
If $c_1=0$ (which implies $r_1=0$) and $f(0)=\infty$, then $\mathcal{R}=\infty$, which means $R_f(\Qbf)=\infty$ for any $\epsilon$-LDP sampling mechanism $\Qbf$.
For the behavior at $\epsilon \rightarrow \infty$ for a fixed $(c_1,c_2)$, if $c_1>0$, then for sufficiently large $\epsilon$, we have $c_1 e^\epsilon \geq c_2$ , so we fall in the aforementioned trivial case that $\mathcal{R}=0$.
If $c_1=0$ and $f(0)<\infty$, then as $\epsilon \rightarrow \infty$, we have $\mathcal{R} \rightarrow 0$, which again corresponds to the non-private case. 

\paragraph*{Remarks on the constant $r_P$}
By the same reason as in the finite space case, the value of $r_P$ may not be unique, but $\Qbf^*_{c_1,c_2,h,\epsilon}$ does not depend on the choice of $r_P$, and $r_P$ can be found by the bisection method with a numerical integration of \eqref{eq:contMech}. 
Note that the continuity of $r \mapsto \int g_r(x) dx , g_r(x)=\clip\left(\frac{1}{r} p(x); bh(x),  be^\epsilon h(x)\right)$, follows from the dominated convergence theorem \cite{Stein15-realAnalysis} since we assume $\int_{\Rbb^n} h(x)dx < \infty$.
The meaning of `Furthermore' part is also similar to the finite space case, that is, the value of $\int g_r(x) dx$ at $r=r_1$ and $r=r_2$ is at least and at most $1$, respectively, so that we can perform the bisection method with initial endpoints $(r_1,r_2)$ (When $r=0$, we define $g_r(x)=be^\epsilon h(x)$ whenever $p(x)>0$ and $g_r(x)= bh(x)$ whenever $p(x)=0$).
A corner case is that when $r_1=0$, the continuity of $r \mapsto \int g_r(x) dx$ does not suffice to guarantee the existence of strictly positive $r$ such that $\int g_r(x) dx=1$.
However, in the proof, we actually show that $\int g_{r_1}(x) dx=1$ implies $\int g_{r}(x) dx=1$ for \emph{every} $r \in (r_1,r_2]$, which especially implies that even when $r_1=0$, there is a \emph{strictly positive} $r$ such that $\int g_r(x) dx=1$. 
This is the reason that we state the strict inequality $r_1<r_P$.

\subsection{Proof sketch of the theorems}\label{sec:proofIdea}
The full proofs of the main theorems, Theorems \ref{thm:optMechFinite} and \ref{thm:optMechCont}, are presented in Appendix \ref{supp:mainThmProof}. In Appendix \ref{supp:mainThmProof}, we present a generalized theorem which includes Theorems \ref{thm:optMechFinite} and \ref{thm:optMechCont} as special cases, where $\Xcal$ can be a general sample space and $\PtildCal$ is similarly defined as in the continuous space case. 
The key idea for proofs and proposed mechanisms is to focus on the behavior of $\Qbf(P)$ when $P$ is in an extreme case in $\PtildCal$. In finite space, point masses are extreme cases, and for continuous space with $\PtildCal=\PtildCal_{c_1,c_2,h}$, the cases that $p(x) \in \{c_1 h(x), c_2 h(x)\}$ for all $x \in \Xcal$ are the extreme cases. As implied by the proof, the worst-case $f$-divergence of the proposed optimal mechanism is attained when $P$ is in the aforementioned extreme cases. Such an approach using extreme case is a frequently used technique in PUT analysis \cite{Kairouz16-ExtremalMechLDP, Holohan17-ExtremePtsLDPPolytope, Pensia23-SimpBinHTLDP, Nam23-OptPrivEstOneBit}.

Our proof consists of two parts, the achievability part and the converse part. The achievability part is to show that the worst-case $f$-divergence $R_f(\Qbf^*)$ of our proposed mechanism $\Qbf^*$ is upper-bounded by the RHS of \eqref{eq:worstDeviationFinCase} or \eqref{eq:worstDeviationBddDensity}. The converse part is to show that $R_f(\Qbf)$ of \emph{any} $\epsilon$-LDP mechanism $\Qbf$ is lower-bounded by the RHS of \eqref{eq:worstDeviationFinCase} or \eqref{eq:worstDeviationBddDensity}. From now, we briefly describe the proof idea of each part. 
{Here, we omit the subscripts $(k,\epsilon)$ or $(c_1,c_2,h,\epsilon)$ for notational convenience.}
{

\paragraph*{Achievability part}
Let $\Mcal = \{\Qbf^*(P): P \in \PtildCal\}$. For  finite space, $\Mcal$ consists of all distributions $Q \in \Pcal([k])$ such that $\frac{1}{e^\epsilon+k-1} \leq Q(x) \leq \frac{e^\epsilon}{e^\epsilon+k-1}$ for every $x \in [k]$. For continuous space, $\Mcal$ consists of all continuous distributions $Q \in \mathcal{C}(\Rbb^n)$ whose pdf $q$ satisfies $bh(x) \leq q(x) \leq be^\epsilon h(x)$ for every $x \in \Rbb^n$.

We construct a mechanism $\Qbf^\dagger$ such that $R_f(\Qbf^\dagger)$ is upper-bounded by the RHS of \eqref{eq:worstDeviationFinCase} or \eqref{eq:worstDeviationBddDensity}. The construction is as follows.
First, we set a reference distribution $\mu \in \Pcal(\Xcal)$ and a constant $\gamma \in [0,1]$ according to {a certain rule specified in Appendix \ref{supp:mainThmProof}.} 
Then, for each given $P$, we generate a private sample by sampling from the original $P$ with probability $\gamma$, and sampling from the reference distribution $\mu$ with probability $1-\gamma$.
In other words, we have $\Qbf^\dagger(P) = \gamma P + (1-\gamma) \mu$.
Our choice of $\mu$ and $\gamma$ makes $\Qbf^\dagger(P) \in \Mcal$ for every $P \in \PtildCal$, which especially implies that $\Qbf^\dagger$ also satisfies $\epsilon$-LDP.
Furthermore, we can find a bound on the ratio of the pmf or pdf for original distribution to that for sampling distribution.
{Then, invoking \cite[Theorem 2.1]{Rukhin97-fDivBddForLRBdd}, 
which bounds $f$-divergences given bounds on the ratio between pmf or pdfs,  we show that $R_f(\Qbf^\dagger)$ is upper-bounded by  the RHS of \eqref{eq:worstDeviationFinCase} or \eqref{eq:worstDeviationBddDensity}.} 

Next, we demonstrate a non-trivial generalization of the main result of \cite{Husain20-LDPSampling} that $\Qbf^*(P)$ is the $f$-divergence projection of $P$ onto $\Mcal$ for $P \in \PtildCal$ and \emph{for every $f$-divergence.}
\begin{proposition}\label{prop:fDivProj}
Assuming the setups of $(\Xcal,\PtildCal,\epsilon)$ in either Theorem \ref{thm:optMechFinite} or \ref{thm:optMechCont}, let $\Qbf^*$ denote the proposed mechanism $\Qbf^*_{k,\epsilon}$ or $\Qbf^*_{c_1,c_2,h,\epsilon}$. Also, let $\Mcal$ be as described above. Then, for every $P \in \PtildCal$ and every $f$-divergence $D_f$, we have $\fDiv{P}{\Qbf^*(P)} = \inf_{Q \in \Mcal} \fDiv{P}{Q}$.
\end{proposition} 
Notice that this proposition differs from \cite{Husain20-LDPSampling} in that it holds for all general $f$-divergences (as opposed to only KL divergence) and general sample spaces, be it discrete or continuous.
This result immediately yields $\fDiv{P}{\Qbf^*(P)} \leq \fDiv{P}{\Qbf^\dagger(P)}$, and hence $R_f(\Qbf^*) \leq R_f(\Qbf^\dagger) \leq (\text{RHS of \eqref{eq:worstDeviationFinCase} or \eqref{eq:worstDeviationBddDensity}})$.
We remark that combining with the converse part (to be described below), it implies that 
the $\Qbf^\dagger$ is also optimal in our minimax sense. 
Nevertheless, $\Qbf^*$ outperforms  $\Qbf^\dagger$ in that $\fDiv{P}{\Qbf^*(P)} \leq \fDiv{P}{\Qbf^\dagger(P)}$ for every $P \in \PtildCal$.
\begin{remark}
    For the finite case, $\mu$ is the uniform distribution over $[k]$ and $\gamma$ is taken in such a way that $\Qbf^\dagger$ satisfies $\epsilon$-LDP tightly. In this case, an alternative way to implement $\Qbf^\dagger$ is as follows. For each given $P$, first we sample from $P$ to get a raw sample and then apply the $k$-ary randomized response \cite{warner1965randomized} to it.
\end{remark}
}

\paragraph*{Converse part}
For each extreme $P$, let {$A_P$ be the ``high probability set'', {defined as $A_P = \{x \in \Xcal : P(x)=1\}$  for finite space and $A_P=\{x \in \Xcal : p(x)=c_2 h(x)\}$ for continuous space.}} Then, using the data processing inequality of $f$-divergence \cite{Liese06-DivInfinStatandIT}, we take a lower bound of $\fDiv{P}{\Qbf(P)}$ 
by the $f$-divergence between the distributions of $\mathbbm{1}_{A_P}(X)$ for $X \sim P$ and that for $X \sim \Qbf(P)$. Such a lower bound becomes a decreasing function of $\Qbf(A_P|P)$ in a certain range.
Then, we seek to find an upper bound on $\inf \Qbf(A_P|P)$ over extreme $P$, which gives a lower bound on $R_f(\Qbf)$. This involves a novel combinatorial argument. We perform a ``packing'' of $u$ copies of $\Xcal$ by $t$ subsets $A_1,\cdots,A_t$ with $A_i = A_{P_i}$ for some extreme $P_i$ and appropriately chosen $t$ and $u$. Then, we decompose the RHS of $u = \sum_{i=1}^{u} \Qbf(\Xcal|P_i)$
by an appropriate partition of $\Xcal$ involving $A_i$'s and use the definition of $\epsilon$-LDP to find an upper bound on $\inf \Qbf(A_P|P)$.

{
\section{Discussions on the Proposed Mechanism}
\subsection[Effect of the r\_P approximation error]{Effect of the $r_P$ approximation error}\label{subsec:effApproxOfConstFactor}
In practice, it may not be possible to find the exact value of $r_P$ such that the sum or integration of the RHS of \eqref{eq:optMechFormFin} or \eqref{eq:contMech} is $1$.
For the case of continuous space as in Theorem \ref{thm:optMechCont}, one way to implement the proposed mechanism  $\Qbf^*_{c_1,c_2,h,\epsilon}$ in practice is as follows. 
First, we fix parameters $\delta_1\in [0, 1)$ and $\delta_2 \geq 0$ that quantify error tolerance. 
For a given $P$, we define $g_r(x)=\clip\left(\frac{1}{r} p(x); bh(x),  be^\epsilon h(x)\right)$ and find $r_P>0$ such that $\int_{\Rbb^n} g_{r_P}(x)dx \in [1-\delta_1, 1+\delta_2]$; 
we delineate a numerical algorithm for this task in Section \ref{subsec:contSpace} based on the bisection method and a numerical integration method.
Then, we get a private sample by sampling from the distribution with pdf $\hat{q}(x)=g_{r_P}(x)/\int_{\Rbb^n} g_{r_P}(x)dx$. 
For the finite case as in Theorem \ref{thm:optMechFinite}, we can implement in the same way, except replacing the integral with the sum.
 
It is important to note that $\hat{q}(x) \in \left[\frac{bh(x)}{1+\delta_2}, \frac{be^\epsilon h(x)}{1-\delta_1}\right]$, indicating that the resulting $\Qbf^*_{k,\epsilon}$ and $\Qbf^*_{c_1,c_2,h,\epsilon}$ satisfy $\left(\epsilon + \log\frac{1+\delta_2}{1-\delta_1}\right)$-LDP as opposed to $\epsilon$-LDP.
Thus, the above implementation yields $\epsilon$-LDP if it is used to implement $\Qbf^*_{k,\epsilon'}$ or $\Qbf^*_{c_1,c_2,h,\epsilon'}$, with $\epsilon' = \epsilon - \log\frac{1+\delta_2}{1-\delta_1}$ and sufficiently small $\delta_1,\delta_2$ such that $\epsilon'>0$.
}
\subsection{Continuity of the proposed mechanism}\label{sec:continuityMech}
In some practical scenarios, the client may not have full access to their distribution $P$. One example is that the client can only access to samples from $P$. In such case, the client may first estimate the true distribution, and then perturb the estimated distribution through the optimal mechanism. The question is how the perturbation using the estimated distribution deviates from that using the true distribution. To answer this, we show that the proposed mechanism satisfies a pointwise Lipschitz property with respect to the total variation distance, and the Lipschitz constant is closely related to the factor $r_P$ we introduce in Theorems \ref{thm:optMechFinite} and \ref{thm:optMechCont}.
\begin{proposition}\label{prop:mechContCoeffBdd}
Assuming the setups of $(\Xcal,\PtildCal,\epsilon)$ in either Theorem \ref{thm:optMechFinite} or \ref{thm:optMechCont}, let $\Qbf^*$ denote the {proposed} mechanism $\Qbf^*_{k,\epsilon}$ or $\Qbf^*_{c_1,c_2,h,\epsilon}$. Then, for any $P,P' \in \PtildCal$, we have
\begin{equation}
    \tvDiv{\Qbf^*(P)}{\Qbf^*(P')} \leq \frac{2}{\max(r_P, r_{P'})} \tvDiv{P}{P'}
\end{equation}
where $r_P>0$ is as in Theorem \ref{thm:optMechFinite} or \ref{thm:optMechCont}.
\end{proposition}
This guarantees that for each given true $P$ and given $\delta>0$, whenever the approximated $P'$ satisfies $\tvDiv{P}{P'} \leq \delta r_P/2$, the perturbed distribution $\Qbf^*(P')$ satisfies $\tvDiv{\Qbf^*(P)}{\Qbf^*(P')} \leq \delta$. In theoretical perspective, this proposition implies that $\Qbf^*$ is continuous when $\PtildCal$ and $\Pcal(\Xcal)$ are endowed with the metric topology from the total variation distance. 
The proof of Proposition \ref{prop:mechContCoeffBdd} is in Appendix~\ref{supp:remainPropStateAndProof}.

\section{Numerical Results}\label{sec:numResult}
In this section, we numerically compare the worst-case $f$-divergence of our proposed mechanism with that of the previously proposed sampling mechanism. To the best of our knowledge, the only work about the private sampling under LDP is \cite{Husain20-LDPSampling}, hence we set the baseline as the mechanism proposed in \cite{Husain20-LDPSampling}. 
In all the cases, we perform the comparison across three canonical $f$-divergences: KL divergence, total variation distance, and squared Hellinger distance, as well as across five values of $\epsilon$: 0.1, 0.5, 1, 2, and 5.

\subsection{Comparison for finite data space}\label{subsec:numResultFinSpace}
\begin{figure}[htbp]
    \centering
    \includegraphics[width=0.8\textwidth]{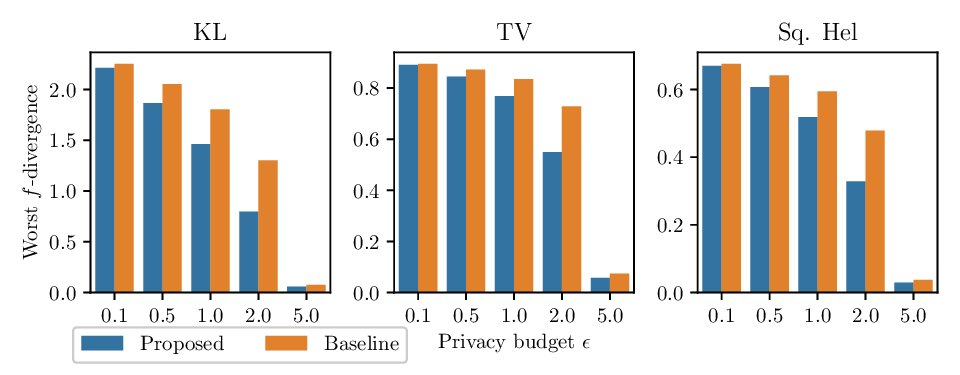}
    \caption{Theoretical worst-case $f$-divergences of proposed and previously proposed baseline mechanisms (with uniform $Q_0$) over finite space ($k=10$)\\
    (Left: KL divergence, Center: Total variation distance, Right: Squared Hellinger distance)}
    \label{fig:expFinSpace}
\end{figure}

In this subsection, we compare the mechanisms in the finite space, $\Xcal = [k]$ and $\PtildCal=\Pcal([k])$. 
As mentioned in Sections \ref{subsec:relWorks} and \ref{subsec:optMechFinSpace}, the baseline mechanism has a hyper-parameter, a reference probability distribution $Q_0 \in \Pcal(\Xcal)$. We set the baseline as a generalized $f$-divergence projection onto the relative mollifier. That is, for each given $f$-divergence, we set the baseline to satisfy $\Qbf(P) \in \argmin_{Q \in \Mcal_{\epsilon,Q_0}} \fDiv{P}{Q}$, where $\Mcal_{\epsilon,Q_0}$ is defined in Section \ref{subsec:relWorks}.
As expected by symmetry, for any $f$, choosing $Q_0$ to be the uniform distribution minimizes the worst-case $f$-divergence $R_f(\Qbf)$ among all choices of $Q_0$ for the baseline. Also, even though we do not obtain the closed-form expression of $\Qbf(x|P)$ for the baseline, we obtain the value of $R_f(\Qbf)$ when $Q_0$ is the uniform distribution. The proof of this fact, together with the precise value of $R_f(\Qbf)$ for uniform $Q_0$, is in Appendix \ref{supp:numResultSetupsFinSpace}. Hence, we always set $Q_0$ to be the uniform distribution in the result about the baseline. Since we have the precise values of $R_f(\Qbf)$ for both our proposed mechanism and the baseline, we plot such values of $R_f(\Qbf)$ in Figure \ref{fig:expFinSpace}. For simplicity, we only provide the plot for $k=10$. More plots for some other $k$'s can be found in Appendix \ref{supp:addExp}. As shown by the figure, the proposed mechanism has lower worst-case $f$-divergence than the baseline for all choices of $f$-divergences and $\epsilon$ in the experiment, with significant gap in medium privacy regime $\epsilon \in [0.5, 2]$.

\subsection{Comparison for 1D Gaussian mixture}\label{subsec:numResult1DGaussMix}
In this subsection, we conduct an experiment to compare the mechanisms when the client distributions are Gaussian mixtures over a real line $\Xcal = \mathbb{R}$, which is an instance of a continuous space case. We consider the case that each client has a Gaussian mixture distribution in $\mathbb{R}$, where each Gaussian has a mean bounded by $1$ and has a unit variance. 
{To avoid arbitrarily large number of Gaussian distributions to be mixed,}
we set an upper bound $K$ of the number of Gaussian distributions to be mixed per client. Also, 
{to make the numerical integration tractable,}
we truncate the domain of the distributions to lie inside an interval $[-4, 4]$. Unlike the finite space case, there is no known closed-form expression of the worst-case $f$-divergence for the mechanism in \cite{Husain20-LDPSampling}. The set of Gaussian mixtures is not exactly of the form $\PtildCal=\PtildCal_{c_1,c_2,h}$, hence our proposed mechanism also does not have a known closed-form expression of the worst-case $f$-divergence. Hence, instead, we compare the mechanisms by an empirical worst-case $f$-divergence.

For an experiment, we randomly construct $N$ Gaussian mixture distributions $P_1,P_2,\cdots,P_{N} \in \PtildCal$, where each $P_j$ is generated independently according to some rules specified in Appendix \ref{supp:numResultSetupsGaussMix}.
After that, we plot the value of the empirical maximum $f$-divergence $\max_{j \in [N]} \fDiv{P_j}{\Qbf(P_j)}$ for the baseline and our proposed $\Qbf$.
For the baseline mechanism, we use MBDE with the same hyperparameter setup as \cite[Section 5]{Husain20-LDPSampling}, {except a slight modification of the reference distribution to consider the truncation of the domain.} The implementation details are provided  in Appendix \ref{supp:numResultSetupsGaussMix}.

In Figure \ref{fig:expGaussMixture1D}, we present the result for $N=100$ and $K=10$. We can see that the proposed mechanism has much lower worst-case $f$-divergence than the baseline for all choices of $f$-divergences and $\epsilon$.

\begin{figure}[htbp]
    \centering
    \includegraphics[width=0.8\textwidth]{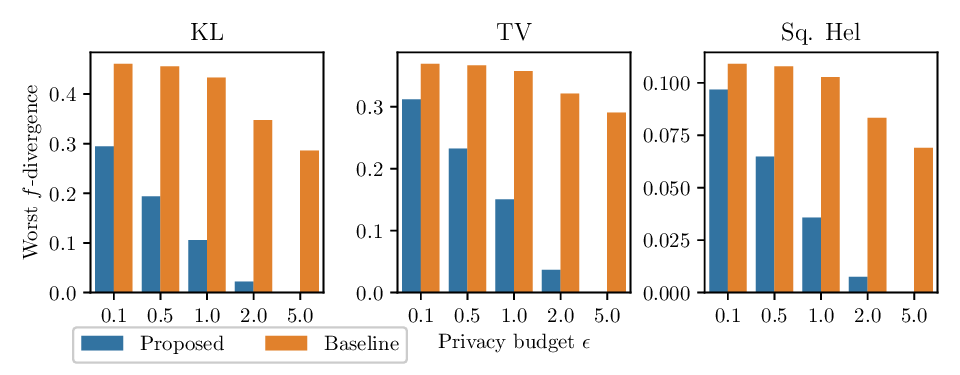}
    \caption{Empirical worst-case $f$-divergences of proposed and baseline mechanisms over 100 experiments of 1D Gaussian mixture\\
    (Left: KL divergence, Center: Total variation distance, Right: Squared Hellinger distance)}
    \label{fig:expGaussMixture1D}
\end{figure}

\section{Conclusion} \label{sec:conclusion}
In this paper, we characterized the optimal privacy-utility trade-off for the private sampling under LDP and found the optimal private sampling mechanism in terms of the minimax $f$-divergence between original and sampling distributions, for both finite and continuous data spaces. Compared to the previous work \cite{Husain20-LDPSampling} based on relative mollifier with arbitrarily chosen reference distribution, our work characterizes PUT without dependency on external information other than the original distribution, and it is shown that the mechanism we found is universally optimal under any $f$-divergence. 

For future works, there may be other $\PtildCal$ and other measures of utility more appropriate to practical scenarios, which are not handled in this paper.
For example, $f$-divergence may be an inappropriate utility loss because it only depends on $\sigma$-algebra structure and does not consider additional information about geometry of $\Xcal$, such as underlying metric on $\Xcal$. Using utility measures involving the geometry, such as Wasserstein distance \cite{Villani09-OTOldNew, Arjovsky17-WGAN, Imola24-metricDPUserLev}, may be more appropriate for some scenarios.
Also, we can consider the Bayesian approach instead of the worst-case approach. 
Furthermore, we only consider the task of releasing a single sample per client in this paper. 
We may also consider the case of releasing multiple samples per client, rather than a single sample. 

The limitations and broader impacts of this work are in Appendices \ref{supp:limitations} and \ref{supp:broaderImpacts}, respectively.



\medskip

{
\small

\bibliographystyle{unsrtnat}
\bibliography{refs}
}

\newpage

\appendices
\section{Assumptions about the Measure Theory}\label{supp:measTheoryAssume}
The appendices assume the familiarity with the basic measure theory and real analysis. We refer the standard textbooks about the measure theory and real analysis, e.g. \cite{Stein15-realAnalysis}.

Throughout the main paper and the appendices, we assume the followings:
\begin{itemize}
    \item For each sample space $\Xcal$, a $\sigma$-algebra on $\Xcal$ is implicitly given. Unless mentioned otherwise,
    \begin{itemize}
        \item the discrete $\sigma$-algebra is given for finite $\Xcal$, and
        \item the Borel $\sigma$-algebra is given for $\Xcal = \mathbb{R}^n$.
    \end{itemize}
    \item A ``subset'' of $\Xcal$ always means a ``measurable subset'', and similarly $A \subset \Xcal$ always means $A$ is a measurable subset.
    \item The ``continuous" distribution precisely means the ``absolutely continuous" distribution (with respect to the Lebesgue measure).
\end{itemize}
   
In the appendices, we also introduce the following notations:
\begin{itemize}
    \item For finite $\Xcal$, $\#$ denotes the counting measure.
    \item For $\Xcal=\Rbb^n$, $m$ denotes the Lebesgue measure.
\end{itemize}

\section{General Definition and More Properties of \texorpdfstring{$f$}{f}-divergences}\label{supp:fDivReview}
In this appendix, we review the general definition and additional properties about the $f$-divergences which are important in our analysis. Let a convex function $f:(0,\infty) \rightarrow \Rbb$ with $f(1)=0$ be given. For $P, Q \in \Pcal(\Xcal)$, not necessarily $P \ll Q$, we first take a dominating measure $\mu$ on $\Xcal$ such that $P,Q \ll \mu$ (e.g., $\mu=P+Q$), and let $p={dP}/{d\mu}$, $q={dQ}/{d\mu}$. The $f$-divergence $\fDiv{P}{Q}$ is defined as
\begin{equation}
    \fDiv{P}{Q} = \int q(x) {f\bracket{\frac{p(x)}{q(x)}}  d\mu (x)}. \label{eq:fDivDefGen}
\end{equation}
We note that \eqref{eq:fDivDefGen} is invariant under the choice of the dominating measure $\mu$. 

Since it is possible that $p(x)=0$ or $q(x)=0$, we need to define what is the value of the expression $qf(p/q)$ for $p=0$ or $q=0$ \cite{Sason18-fDiv, SasonVerdu16-fDivIneqs}.
It is well-known that any convex function $f:(0,1) \rightarrow \Rbb$ is continuous, and the function 
\begin{align}
    f^\star(x)=xf(1/x)
\end{align}
is also convex on $(0,1)$. Furthermore, any convex function $f:(0,1) \rightarrow \Rbb$ with $f(1)=0$ has a limit $\lim_{x \rightarrow 0+}f(x)$ in $\Rbb \cup \{+\infty\}$. Hence, we have the continuous extension $f:[0,\infty) \rightarrow \Rbb \cup \{+\infty\}$ by setting $f(0)=\lim_{x \rightarrow 0+}f(x)$, which is proper convex and continuous. By the same way, we have the continuous extension $f^\star : [0,\infty) \rightarrow \Rbb \cup \{+\infty\}$. Using these extensions, we define $0 f(0/0)=0$, $q f(0/q)=qf(0)$ for $q>0$, and $0 f(p/0) = p f^\star(0)$ for $p>0$. Especially, if $f^\star(0)=\infty$, then $\fDiv{P}{Q} = \infty$ whenever $P \ll Q$ does not hold. (This is the case for KL divergence and $\chi^2$ divergence for examples) Similarly, if $f(0)=\infty$, then $\fDiv{P}{Q} = \infty$ whenever $Q \ll P$ does not hold. Also, the maximum value of the $f$-divergence $M_f$ presented in \eqref{eq:fDivMutSing} can be written as $M_f=f(0)+f^\star (0)$.

The following additional properties of $f$-divergences are important in our analysis. \cite{Liese06-DivInfinStatandIT}
\begin{theorem}\label{thm:fDivConv}
Any $f$-divergence is jointly convex, that is for any $P_1,P_2,Q_1,Q_2 \in \Pcal(\Xcal)$ and $0 \leq \lambda \leq 1$, we have $\fDiv{\lambda P_1 + (1-\lambda) P_2}{\lambda Q_1 + (1-\lambda) Q_2} \leq \lambda \fDiv{P_1}{Q_1} + (1-\lambda)\fDiv{P_2}{Q_2}$.    
\end{theorem}
\begin{theorem}[Data-Processing Inequality]\label{thm:fDivDPI}
Let $M$ be a conditional distribution (Markov kernel) from $\mathcal{X}$ to $\mathcal{Y}$. For given $P_1, P_2 \in \Pcal(\Xcal)$, let $Q_1, Q_2 \in \Pcal(\mathcal{Y})$ be the push-forward measure of $P_1,P_2$ through $M$, respectively. Then for any $f$-divergence, we have $\fDiv{P_1}{P_2} \geq \fDiv{Q_1}{Q_2}$.
\end{theorem}

Also, we present several equivalent expressions for the total variation distance \cite{Sason18-fDiv, SasonVerdu16-fDivIneqs}, which are used in Appendix \ref{supp:remainPropStateAndProof}. In below expressions, the assumption is that $P,Q \ll \mu$ for some dominating measure $\mu$ with $p=dP/d\mu$ and $q=dQ/d\mu$.
\begin{align}
    \tvDiv{P}{Q} &= \frac{1}{2}\int |p(x)-q(x)| d\mu(x)\\
              &= \int_{x:p(x) \leq q(x)} (q(x)-p(x)) d\mu(x)\\
              &= \int_{x:p(x) \geq q(x)} (p(x)-q(x)) d\mu(x). \label{eq:TVrepPosPart}
\end{align}
We introduce one more notation. For $\lambda_1, \lambda_2 \in [0,1]$, let $\fDivBin{\lambda_1}{\lambda_2}$ denotes the $f$-divergence between Bernoulli distributions with $\mathrm{Pr}(1)=\lambda_1$ and $\lambda_2$, respectively. That is,
\begin{align}
    \fDivBin{\lambda_1}{\lambda_2} = \lambda_2 f\left(\frac{\lambda_1}{\lambda_2}\right) + (1-\lambda_2)f\left(\frac{1-\lambda_1}{1-\lambda_2} \right).
\end{align}
We should note the following facts:
\begin{enumerate}
    \item By joint convexity of the $f$-divergence and continuity of $f$, $\fDivBin{\lambda_1}{\lambda_2}$ is continuous and jointly convex in $(\lambda_1,\lambda_2)$. (But $\fDivBin{\lambda_1}{\lambda_2}$ may be extended real-valued)
    \item For a fixed $\lambda_1$, $\fDivBin{\lambda_1}{\lambda_2}$ attains a global minimum $0$ at $\lambda_2=\lambda_1$. Together with convexity, we derive that $\fDivBin{\lambda_1}{\lambda_2}$ is decreasing in $\lambda_2 \in [0,\lambda_1]$ and increasing in $\lambda_2 \in [\lambda_1,1]$, respectively.
\end{enumerate}

\section{Generalized Main Theorem and Proof}\label{supp:mainThmProof}
In this appendix, we present the formal proofs of the main theorems, Theorems \ref{thm:optMechFinite} and \ref{thm:optMechCont}. As mentioned in Section \ref{sec:proofIdea}, we first state the generalized theorem with its proof, and later we show how this generalized theorem includes the main theorems as special cases.

\subsection{Statement of the generalized main theorem}
First, let us define the general setup we consider. Let a (general) sample space $\Xcal$ be given. For a positive measure $\mu$ on $\Xcal$ such that $\mu(\Xcal) < \infty$ and $c_2>c_1 \geq 0$, let us define
\begin{align}
    \PtildCal_{c_1, c_2, \mu} := \{P \in \Pcal(\Xcal): P \ll \mu, \quad c_1 \leq dP/d\mu \leq c_2 \quad \mu\text{-a.e.}\}.
\end{align}
We generally consider the case that $\PtildCal = \PtildCal_{c_1, c_2, \mu}$ for some $c_1,c_2,\mu$. For example, for the setup of Section \ref{subsec:contSpace}, we have $\PtildCal_{c_1,c_2,h} = \PtildCal_{c_1,c_2,\mu}$, where $\mu$ is a positive measure with $\mu \ll m$ and $d\mu/dm = h$. For the setup of Section \ref{subsec:optMechFinSpace}, we have $\Pcal([k]) = \PtildCal_{0, 1, \#}$. 
By the same reason as in Section \ref{subsec:contSpace}, we may impose the following normalization condition on $c_1,c_2,\mu,\epsilon$:
\begin{align}
    \mu(\Xcal) = 1, \quad c_1 < 1 < c_2, \quad c_2 > c_1 e^\epsilon. \label{eq:normalizationCondGeneral}
\end{align}
Note that $\mu(\Xcal)=1$ means that $\mu$ is a probability measure, that is $\mu \in \Pcal(\Xcal)$. Also, note that for $\Xcal=[k]$, we can write in normalized form as $\Pcal([k]) = \PtildCal_{0, k, \mu_k}$, where $\mu_k = \frac{1}{k}\#$ is the uniform distribution on $[k]$.

First, let us define the proposed mechanism, together with the related constants as the same as Theorem \ref{thm:optMechCont}. For the ease of proof, we introduce an additional constant $\alpha$ over Theorem \ref{thm:optMechCont}.
\begin{definition}\label{def:mechDefGenSetup}
    Let $c_1,c_2,\mu,\epsilon$ satisfying the normalization condition \eqref{eq:normalizationCondGeneral} be given, and let $\PtildCal = \PtildCal_{c_1,c_2,\mu}$. 
    First, define the following constants determined by $c_1,c_2,\epsilon$:
    \begin{align}
        \alpha &= \frac{1-c_1}{c_2 - c_1} \label{eq:alphaDef},\\
        b &= \frac{c_2 - c_1}{(e^\epsilon-1)(1-c_1)+c_2-c_1} = \frac{1}{\alpha e^\epsilon + 1 - \alpha},\\
        r_1 &= \frac{c_1}{b} = \left(\frac{(e^\epsilon-1)(1-c_1)+c_2-c_1}{c_2 - c_1}\right) c_1 = c_1 (\alpha e^\epsilon + 1 - \alpha),\\
        r_2 &= \frac{c_2}{b e^\epsilon} = \left(\frac{(e^\epsilon-1)(1-c_1)+c_2-c_1}{c_2 - c_1}\right) \frac{c_2}{e^\epsilon} = \frac{c_2}{e^\epsilon}(\alpha e^\epsilon + 1 - \alpha).
    \end{align}
    Also, define a mechanism $\Qbf^*_{c_1,c_2,\mu,\epsilon} \in  \mathcal{Q}_{\Xcal,\PtildCal, \epsilon}$ as follows:

    \emph{For each $P \in \PtildCal$, $\Qbf^*_{c_1,c_2,\mu,\epsilon}(P)=:Q$ is defined as a probability measure such that $Q \ll \mu$ and
    \begin{equation}
        \frac{dQ}{d\mu}(x) = \clip\left(\frac{1}{r_P} \frac{dP}{d\mu}(x); b,  be^\epsilon \right), \label{eq:generalMech}
    \end{equation}
    where $r_P>0$ is a constant depending on $P$ so that $\int \frac{dQ}{d\mu}d\mu(x) = 1$.}
    Furthermore, let $\Mcal_{c_1,c_2,\mu ,\epsilon} = \PtildCal_{b,be^\epsilon,\mu}$, so that $\Qbf^*_{c_1,c_2,\mu,\epsilon}(P) \in \Mcal_{c_1,c_2,\mu ,\epsilon}$ for every $P \in \PtildCal$.
\end{definition}
We should note that $1=c_2 \alpha + c_1 (1-\alpha) = be^\epsilon \alpha + b(1-\alpha)$. Also, $\Qbf^*_{c_1,c_2,\mu,\epsilon}$ clearly satisfies $\epsilon$-LDP.

By the same reason as in Sections \ref{subsec:optMechFinSpace} and \ref{subsec:contSpace}, the values of $r_P$ may not be unique, but the mechanism $\Qbf^*_{c_1,c_2,\mu,\epsilon}$ does not depend on the choice of $r_P$. Furthermore, we show that the value of $\int \clip\left(\frac{1}{r} \frac{dP}{d\mu}(x); b,  be^\epsilon \right) d\mu(x)$ at $r=r_1$ and $r=r_2$ are at least and at most $1$, respectively. 
\begin{proposition}\label{prop:normFactorRange}
    Let $c_1,c_2,\mu,\epsilon,\PtildCal,\alpha,b,r_1,r_2$ be as in Definition \ref{def:mechDefGenSetup}. Then, for any $P \in \PtildCal$, we have
    \begin{align}
        \int \clip\left(\frac{1}{r_1} \frac{dP}{d\mu}(x); b,  be^\epsilon \right) d\mu(x) \geq 1 \geq \int \clip\left(\frac{1}{r_2} \frac{dP}{d\mu}(x); b,  be^\epsilon \right) d\mu(x),
    \end{align}
    where, in the case of $r_1=0$, we define
    \begin{align}
        \clip\left(\frac{1}{r_1} \frac{dP}{d\mu}(x); b,  be^\epsilon \right) = \begin{cases}
            b, & \text{ if }\frac{dP}{d\mu}(x)=0\\
            be^\epsilon, & \text{ otherwise}
        \end{cases}.
    \end{align}
    Furthermore, if $\int \clip\left(\frac{1}{r_1} \frac{dP}{d\mu}(x); b,  be^\epsilon \right) d\mu(x)=1$, then $\int \clip\left(\frac{1}{r} \frac{dP}{d\mu}(x); b,  be^\epsilon \right) d\mu(x)=1$ for every $r \in [r_1,r_2]$.
\end{proposition}
This proposition, together with the fact that $r \mapsto \int \clip\left(\frac{1}{r} \frac{dP}{d\mu}(x); b,  be^\epsilon \right) d\mu(x)$ is continuous and monotone decreasing, implies that $r_P$ can be chosen such that $r_1 < r_P \leq r_2$, as stated in Theorem \ref{thm:optMechCont}. (Again, the continuity is from $\mu(\Xcal) < \infty$ and the dominated convergence theorem)

Also, we should remark that $0 < \alpha < 1$ and $c_1 < b < 1 < be^\epsilon < c_2$. The first one easily follows from $c_1<1<c_2$ and the definition of $\alpha$. For the second one, first we have $1 < \alpha e^\epsilon + 1 - \alpha < e^\epsilon$, as $\alpha e^\epsilon + 1 - \alpha$ is a propoer convex combination of $1$ and $\epsilon$. This directly implies that $b<1<be^\epsilon$. Next, by calculations, we can observe that
\begin{align}
    c_1 ((e^\epsilon-1)(1-c_1)+c_2-c_1) - (c_2-c_1) = (1-c_1)(c_1 e^\epsilon - c_2) < 0, \label{ineq:c1AndbComp}
\end{align}
which implies $c_1 < b$, and
\begin{align}
    c_2((e^\epsilon-1)(1-c_1)+c_2-c_1) -e^\epsilon (c_2 - c_1) = (c_2 - 1)(c_2 - e^\epsilon c_1) > 0, \label{ineq:c2AndbComp}
\end{align}
which implies $be^\epsilon < c_2$. Especially, these inequalities imply that $0 \leq r_1 < 1 < r_2$.

Now, we show that under a mild `decomposability' condition, the proposed mechanism $\Qbf^*_{c_1,c_2,\mu,\epsilon}$ is universally optimal under any $f$-divergences for $(\Xcal, \PtildCal=\PtildCal_{c_1,c_2,\mu}, \epsilon)$. After that, we show that such a mild condition holds for the setups of both of the main theorems, which finishes the proofs of the main theorems. 

First, let us state the `decomposability' condition. This condition is a formal definition of the concept of ``packing of $u$ copies of $\Xcal$ by $t$ subsets $A_1,\cdots,A_t$'', which is briefly mentioned in Section \ref{sec:proofIdea},
\begin{definition}
Let $\alpha \in (0,1)$ and $t,u \in \Nbb$, $t>u$. We say that a probability measure $\mu \in \Pcal(\Xcal)$ is \textbf{$(\alpha,t,u)$-decomposable} if there exist $t$  subsets $A_1,A_2,\cdots,A_t \subset \Xcal$ such that $\mu(A_i)=\alpha$ for all $i \in [t]$, and for every $x \in \Xcal$, we have $|\bracketSet{i \in [t] : x \in A_i}| \leq u$.

We say that $\mu \in \Pcal(\Xcal)$ is \textbf{$\alpha$-decomposable} if for any $\delta > 0$, there exists $t,u \in \Nbb$, $t>u$, such that $\alpha \leq u/t < \alpha + \delta$, and $\mu$ is $(\alpha,t,u)$-decomposable.

We remark that $(\alpha,t,1)$-deomposability means that there are $t$ disjoint  subsets $B_1,B_2,\cdots,B_{t}$ such that $\mu(B_i)=\alpha$ for each $i \in [t]$. Also, if $\alpha$ is a rational number with $\alpha=u/t$, $u,t \in \Nbb$, and $\mu$ is $(\alpha,t,u)$-decomposable, then $\mu$ is $\alpha$-decomposable.
\end{definition}

Then, we state the generalized theorem.
\begin{theorem}\label{thm:mainThmGen}
    Let $c_1,c_2,\mu,\epsilon,\PtildCal,\alpha,b,r_1,r_2$ be as in Definition \ref{def:mechDefGenSetup}. If $\mu$ is $\alpha$-decomposable, then for any $f$-divergences $D_f$, we have
    \begin{align}
    \mathcal{R}(\Xcal,\PtildCal, \epsilon, f) = \frac{1-r_1}{r_2-r_1} f(r_2) + \frac{r_2 - 1}{r_2 - r_1} f(r_1),
     \label{eq:worstDeviationGen}
    \end{align}
    and furthermore, the mechanism $\Qbf^*_{c_1,c_2,\mu,\epsilon}$ as in Definition \ref{def:mechDefGenSetup} is optimal for $(\Xcal, \PtildCal, \epsilon)$ under any $f$-divergences $D_f$.
\end{theorem}

As guided in Section \ref{sec:proofIdea}, the proof of this theorem is broken into two parts, the achievability part and the converse part.
\begin{proposition}[Achievability part]\label{prop:achievability}
    Let $c_1,c_2,\mu,\epsilon,\PtildCal,\alpha,b,r_1,r_2$ be as in Definition \ref{def:mechDefGenSetup}. Then, for any $f$-divergences $D_f$, we have
    \begin{align}
        R_f(\Qbf^*_{c_1,c_2,\mu,\epsilon}) \leq \frac{1-r_1}{r_2-r_1} f(r_2) + \frac{r_2 - 1}{r_2 - r_1} f(r_1).
    \end{align}
    (Here, we do not need to assume that $\mu$ is $\alpha$-decomposable)
\end{proposition}

\begin{proposition}[Converse part]\label{prop:converse}
    Let $c_1,c_2,\mu,\epsilon,\PtildCal,\alpha,b,r_1,r_2$ be as in Definition \ref{def:mechDefGenSetup}, and suppose that $\mu$ is $\alpha$-decomposable. Then for any $f$-divergences $D_f$ and for any $\epsilon$-LDP mechanism $\Qbf \in \mathcal{Q}_{\Xcal,\PtildCal, \epsilon}$, we have
    \begin{align}
        R_f(\Qbf) \geq \frac{1-r_1}{r_2-r_1} f(r_2) + \frac{r_2 - 1}{r_2 - r_1} f(r_1).
    \end{align}
\end{proposition}

The remaining of this appendix is organized as follows. 
We first present the proofs of Propositions \ref{prop:achievability} and \ref{prop:converse} in Appendices \ref{subsec:AchievProof} and \ref{subsec:ConverseProof}, in which we grant Proposition \ref{prop:normFactorRange} and some intermediate lemmas. 
After that, in Appendix \ref{subsec:deduceMainThm}, we show that Theorem \ref{thm:mainThmGen} contains main theorems, Theorems \ref{thm:optMechFinite} and \ref{thm:optMechCont}, as special cases.
Finally, Appendix \ref{subsec:proofPropNormFactorRange} presents the proof of Proposition \ref{prop:normFactorRange}, and Appendices \ref{subsec:proofLemDensityRatioBddToDivBdd} and \ref{subsec:proofLemDecompMinProbBound} prove intermediate lemmas.

\subsection{Proof of achievability part (Proposition \ref{prop:achievability})}\label{subsec:AchievProof}
{
As mentioned in Section \ref{sec:proofIdea}, the proof of the achievability part consists of two steps: first presenting an alternative mechanism and then proving that $\Qbf^*_{c_1,c_2,\mu,\epsilon}$ performs the $f$-divergence projection for any general $f$-divergence.
\subsubsection{An alternative mechanism}
Let $\Qbf_{c_1,c_2,\mu,\epsilon}^\dagger \in \mathcal{Q}_{\Xcal,\PtildCal, \epsilon}$ be a mechanism defined as follows:
\begin{align}
    \Qbf_{c_1,c_2,\mu,\epsilon}^\dagger (P) = \gamma P + (1-\gamma)\mu,
\end{align}
where
\begin{align}
    \gamma = \frac{e^\epsilon-1}{(e^\epsilon-1)(1-c_1)+c_2-c_1}.
\end{align}
Since $c_2>e^\epsilon c_1$, it follows that $0<\gamma<1$. Also, a direct computation shows that
\begin{align}
    b &= \gamma c_1 + (1-\gamma), \label{eq:gamma1}\\
    be^\epsilon &= \gamma c_2 + (1-\gamma). \label{eq:gamma2}
\end{align}
Notice that since $c_1 \leq \frac{dP}{d\mu} \leq c_2$ and $\frac{d\Qbf_{c_1,c_2,\mu,\epsilon}^\dagger (P)}{d\mu} = \gamma \frac{dP}{d\mu} + (1-\gamma)$, we have 
\begin{align}
    \frac{d\Qbf_{c_1,c_2,\mu,\epsilon}^\dagger (P)}{d\mu}(x)  &\geq \gamma c_1 + (1-\gamma) = b, \\
    \frac{d\Qbf_{c_1,c_2,\mu,\epsilon}^\dagger (P)}{d\mu}(x)   &\leq \gamma c_2 + (1-\gamma) = be^\epsilon.
\end{align}
Hence, $\Qbf_{c_1,c_2,\mu,\epsilon}^\dagger (P) \in \Mcal_{c_1,c_2,\mu ,\epsilon}$ for every $P \in \PtildCal$, implying that $\Qbf_{c_1,c_2,\mu,\epsilon}^\dagger$ also satisfies $\epsilon$-LDP.

Now, we show that 
\begin{align}
    R_f(\Qbf^\dagger_{c_1,c_2,\mu,\epsilon}) \leq \frac{1-r_1}{r_2-r_1} f(r_2) + \frac{r_2 - 1}{r_2 - r_1} f(r_1). \label{eq:altMechAchiev}
\end{align}
To this end, we need to show that for each $P \in \PtildCal$, we have
\begin{align}
    \fDiv{P}{\Qbf^\dagger_{c_1,c_2,\mu,\epsilon}(P)} \leq \frac{1-r_1}{r_2-r_1} f(r_2) + \frac{r_2 - 1}{r_2 - r_1} f(r_1). \label{eq:altMechAchievPoint}
\end{align}
Fix $P \in \PtildCal$.
Let $p=dP/d\mu$ and $q=d\Qbf^\dagger_{c_1,c_2,\mu,\epsilon}(P)/d\mu$. 
First, we claim that
\begin{equation}
    r_1 \leq \frac{p(x)}{q(x)} \leq r_2
\end{equation}
for $\mu$-almost every $x \in \Xcal$. We have
\begin{align}
    \frac{p(x)}{q(x)} 
    &= \frac{p(x)}{\gamma p(x) + (1-\gamma)}\\
    &= \frac{1}{\gamma}\left(1-\frac{1-\gamma}{\gamma p(x) + (1-\gamma)}\right),
\end{align}
which is increasing in $p(x) \geq 0$. Since $c_1 \leq p(x) \leq c_2$, we have
\begin{align}
    \frac{c_1}{\gamma c_1 + (1-\gamma)} \leq \frac{p(x)}{q(x)} \leq \frac{c_2}{\gamma c_2 + (1-\gamma)}
\end{align}
for $\mu$-almost every $x \in \Xcal$.
From \eqref{eq:gamma1}, \eqref{eq:gamma2}, and Definition \ref{def:mechDefGenSetup}, we conclude that
\begin{equation}
    r_1 \leq \frac{p(x)}{q(x)} \leq r_2,
\end{equation}
which proves the claim. For the remaining of the proof, we need the following lemma.
\begin{lemma}[\cite{Rukhin97-fDivBddForLRBdd}, Theorem 2.1]\label{lem:densityRatioBddToDivBdd}
Let $P,Q \in \Pcal(\Xcal)$. Suppose that $P,Q \ll \mu$ for some reference measure $\mu$ on $\Xcal$, and there exist $r_1,r_2 \in \mathbb{R}$ with $0 \leq r_1 < 1 < r_2$ such that the densities $p=\frac{dP}{d\mu}, q=\frac{dQ}{d\mu}$ satisfy $q(x)>0$ and $r_1 \leq \frac{p(x)}{q(x)} \leq r_2$ for $\mu$-almost every $x \in \mathcal{X}$. Then for any $f$-divergence $D_f$, we have
\begin{align}
    \fDiv{P}{Q} \leq \frac{1-r_1}{r_2-r_1}f(r_2) + \frac{r_2-1}{r_2-r_1}f(r_1).
\end{align}
\end{lemma}
This lemma directly implies \eqref{eq:altMechAchievPoint}, and consequently \eqref{eq:altMechAchiev}.

\subsubsection[f-divergence projection of the proposed mechanism]{$f$-divergence projection of the proposed mechanism}
Next, we prove that
$\Qbf_{c_1,c_2,\mu,\epsilon}^*(P)$ is the projection of $P$ onto $\Mcal_{c_1,c_2,\mu ,\epsilon}$ for \emph{every $f$-divergence}.
Proposition \ref{prop:fDivProj} can be  stated in a more general way as follows.  
\begin{proposition}\label{prop:fDivProjGen}
    For any $P \in \PtildCal$ and any $f$-divergences $D_f$, we have
    \begin{align}
    \fDiv{P}{\Qbf_{c_1,c_2,\mu,\epsilon}^* (P)} = \inf_{Q \in \Mcal_{c_1,c_2,\mu ,\epsilon}} \fDiv{P}{Q}.
\end{align}
\end{proposition}
This proposition implies that 
\begin{align}
    \fDiv{P}{\Qbf_{c_1,c_2,\mu,\epsilon}^*(P)} \leq \fDiv{P}{\Qbf_{c_1,c_2,\mu,\epsilon}^\dagger(P)},
\end{align}
for every $P \in \PtildCal$, and hence
\begin{align}
    R_f(\Qbf_{c_1,c_2,\mu,\epsilon}^*) \leq R_f(\Qbf_{c_1,c_2,\mu,\epsilon}^\dagger) \leq \frac{1-r_1}{r_2-r_1}f(r_2) + \frac{r_2-1}{r_2-r_1}f(r_1),
\end{align}
which completes the proof of the achievability part.

Next, we prove Proposition \ref{prop:fDivProjGen}. 
Fix $P \in \PtildCal$ and $p=dP/d\mu$. If $f(0)=\infty$ and $\mu(\{x:p(x)=0\})>0$, then $\fDiv{P}{Q}=\infty$ for every ${Q \in \Mcal_{c_1,c_2,\mu ,\epsilon}}$, thus the proposition holds trivially. Consequently, we assume either $f(0)<\infty$ or $\mu(\{x:p(x)=0\})=0$.

The optimization problem $\inf_{Q \in \Mcal_{c_1,c_2,\mu ,\epsilon}} \fDiv{P}{Q}$ can be cast as the following
\begin{align}
    \inf_{q:\Xcal \rightarrow (0,\infty)} \int q(x)f\left(\frac{p(x)}{q(x)}\right)d\mu(x)
\end{align}
\begin{align}
    \text{such that } \,\,\,\,
     q(x) &\geq b, \quad \forall x,\\
     q(x) &\leq be^\epsilon,  \quad \forall x,\\
    \int q(x)d\mu(x) &= 1,
\end{align}
where $q=dQ/d\mu$.
Our goal is to show that $q^*=d\Qbf^*_{c_1,c_2,\mu,\epsilon}(P)/d\mu$ is an optimal solution of the above optimization problem.

\paragraph*{Motivation for the optimality proof}
To provide the motivation for the proof of the optimality of $q^*$ for the above optimization problem,
we first consider the following analogous finite-dimensional optimization problem
\begin{align}
    \inf_{q \in (0,\infty)^n} \sum_{i=1}^{n} q_i f\left(\frac{p_i}{q_i}\right)
\end{align}
\begin{align}
    \text{such that } \,\,\,\,
     q_i &\geq b, \label{eq:toyPrimalFeas1}\\
     q_i &\leq be^\epsilon, \label{eq:toyPrimalFeas2}\\
    \sum_{i=1}^{n} q_i &= 1. \label{eq:toyPrimalFeas3}
\end{align}
for convex function $f:(0,\infty) \rightarrow \mathbb{R}$ and $p_i > 0$.
The convexity of $f^\star (x) = xf(1/x)$ (see Appendix \ref{supp:fDivReview}) implies that the above is a convex optimization problem.
For simplicity, we assume that $f^\star$ is differentiable. However, note that this assumption is only for simplifying the motivation for the proof, and the result holds for general $f$ or $f^\star$. 

We formulate the Lagrangian for the above optimization as
\begin{align}
    \mathcal{L}(q,\phi,\psi, \nu) = \sum_{i=1}^n \left(p_i f^\star (q_i/p_i) + \phi_i(b-q_i) + \psi_i(q_i - be^\epsilon) \right) + \nu \left(1- \sum_{i=1}^{n} q_i \right)
\end{align}
with dual variables $\phi,\psi \in [0,\infty)^n$ and $\nu \in \mathbb{R}$.

The Karush-Kuhn-Tucker (KKT) condition yields:
\begin{align}
    (f^\star)'(q_i/p_i) - \phi_i + \psi_i - \nu = 0, \quad \forall i, \label{eq:toyKKT1}\\
    \eqref{eq:toyPrimalFeas1},  \eqref{eq:toyPrimalFeas2}, \eqref{eq:toyPrimalFeas3}, \label{eq:toyKKT2}\\
    \phi_i, \psi_i \geq 0, \quad \forall i,\label{eq:toyKKT3}\\
    \phi_i(q_i-b) = \psi_i(be^\epsilon-q_i)=0, \quad \forall i. \label{eq:toyKKT4}
\end{align}
Now, suppose that there is a feasible point $q^* \in (0,\infty)^n$ satisfying \eqref{eq:toyPrimalFeas1}, \eqref{eq:toyPrimalFeas2}, and \eqref{eq:toyPrimalFeas3} such that $q^* = \clip(p_i/r;b,be^\epsilon)$ for some $r>0$. We show that $q^*$ satisfies the KKT condition for some feasible dual variables $(\phi,\psi,\nu)$.

For $(q^*,\phi,\psi,\nu)$ to satisfy the KKT condition, the following should hold:
\begin{itemize}
    \item If $b<p_i/r<be^\epsilon$, then we have $q^*_i = p_i/r$. From \eqref{eq:toyKKT1} and \eqref{eq:toyKKT4}, we must have $\phi_i=\psi_i=0$ and $\nu = (f^\star)'(1/r)$.
    \item If $p_i/r \leq b$, then $q^*_i=b$. Then, $\psi_i=0$ and $\phi_i=(f^\star)'(b/p_i)-\nu$.
    \item If $p_i/r \geq be^\epsilon$, then $q^*_i = be^\epsilon$. Then $\phi_i=0$ and $\psi_i = \nu - (f^\star)'(be^\epsilon/p_i)$.
\end{itemize}
Now, since $f^\star$ is convex, $(f^\star)'$ is monotonically increasing. Hence, the following should be satisfied: 
\begin{itemize}
    \item If $p_i/r \leq b$, then $(f^\star)'(b/p_i)-(f^\star)'(1/r) \geq 0$, and
    \item If $p_i/r \geq be^\epsilon$, then $(f^\star)'(1/r) - (f^\star)'(be^\epsilon/p_i) \geq 0$.
\end{itemize}
It can thus be verified that $(q^*,\phi,\psi,\nu)$ satisfies the KKT condition with
\begin{itemize}
    \item $\nu = (f^\star)'(1/r)$,
    \item $\phi_i = \begin{cases}
        (f^\star)'(b/p_i)-(f^\star)'(1/r), & \text{ if } p_i/r \leq b,\\
        0, & \text{ otherwise,}
    \end{cases}$
    \item $\psi_i = \begin{cases}
        (f^\star)'(1/r) - (f^\star)'(be^\epsilon/p_i), & \text{ if } p_i/r \geq be^\epsilon,\\
        0, & \text{ otherwise.}
    \end{cases}$
\end{itemize}

\paragraph*{Optimality proof}
Next, we present a proof for the optimality of $q^*$ for the optimization problem
\begin{align}
    \inf_{q:\Xcal \rightarrow (0,\infty)} \int q(x)f\left(\frac{p(x)}{q(x)}\right)d\mu(x)
\end{align}
\begin{align}
    \text{such that } \,\,\,\,
     q(x) &\geq b, \quad \forall x, \label{eq:origOptProbCond1}\\
     q(x) &\leq be^\epsilon,  \quad \forall x, \label{eq:origOptProbCond2}\\
    \int q(x)d\mu(x) &= 1. \label{eq:origOptProbCond3}
\end{align}
Here, note that we do not assume $f$ is differentiable. To this goal, we first review the following basic facts about a general convex function $f:(0,\infty) \rightarrow \mathbb{R}$, which \emph{may not be differentiable} \cite{Boyd04-convOptBook, Rockafellar70-ConvAnalysisBook}:
    \begin{itemize}
        \item The left derivative $f_-'(x):=\lim_{h \rightarrow 0-} \frac{f(x+h)-f(x)}{h}$ and the right derivative $f_+'(x):=\lim_{h \rightarrow 0+} \frac{f(x+h)-f(x)}{h}$ exist and finite for every $x \in (0,\infty)$, regardless of whether $f$ is differentiable or not.
        \item For every $0<x<y$, we have $f_-'(x) \leq f_+'(x) \leq f_-'(y) \leq f_+'(y)$.
        \item For every $x,y \in (0,\infty)$ and any $g \in [f_-'(x),f_+'(x)]$, we have
        \begin{align}
            f(y) \geq f(x) + g(y-x). \label{eq:firstOrderLB}
        \end{align}
        By continuous extension, this holds for $y=0$ also. That is, $f(0) \geq f(x)-gx$ for every $x \in (0,\infty)$ and $g \in [f_-'(x),f_+'(x)]$.
    \end{itemize}

Let $q:\Xcal \rightarrow (0,\infty)$ be any feasible function satisfying \eqref{eq:origOptProbCond1} - \eqref{eq:origOptProbCond3}. 
Recall that $f^\star(x):=xf(1/x)$ is convex, and we can express $q(x)f(p(x)/q(x))=p(x)f^\star(q(x)/p(x))$ whenever $p(x) \neq 0$. Also, whenever $p(x) \neq 0$, we bound $f^\star(q(x)/p(x))$ by the linear approximation of $f^\star$ at $q^*(x)/p(x)$ using \eqref{eq:firstOrderLB}, as follows:
\begin{align}
    f^\star\left(\frac{q(x)}{p(x)}\right) \geq f^\star\left(\frac{q^*(x)}{p(x)}\right) + (f^\star)_+'\left(\frac{q^*(x)}{p(x)}\right)\left[\frac{q(x)}{p(x)}-\frac{q^*(x)}{p(x)}\right].
\end{align}
Hence, we have
\begin{align}
    q(x)f\left(\frac{p(x)}{q(x)}\right) = p(x) f^\star\left(\frac{q(x)}{p(x)}\right) \geq q(x) \zeta(x) + \xi(x),
\end{align}
where
\begin{align}
    \zeta(x) &= (f^\star)_+'\left(\frac{q^*(x)}{p(x)}\right),\\
    \xi(x) &= p(x) f^\star\left(\frac{q^*(x)}{p(x)}\right) - q^*(x) (f^\star)_+'\left(\frac{q^*(x)}{p(x)}\right).
\end{align}
Also, whenever $p(x)=0$, we have $q(x)f(p(x)/q(x))=q(x)p(0)$. Hence, we set $\zeta(x)=p(0)$ and $\xi(x)=0$ when $p(x)=0$, so that
\begin{align}
    q(x)f\left(\frac{p(x)}{q(x)}\right) \geq q(x) \zeta(x) + \xi(x)
\end{align}
holds for all $x \in \Xcal$. We note that $\zeta(x)$ and $\xi(x)$ does not depend on the choice of $q(x)$. Also, the equality holds if $q(x)=q^*(x)$.

Next, as an analogous to $\nu$ in the above motivation for the proof,
let $\nu=(f^\star)_+'(1/r_P)$. Since $\int q(x)d\mu(x)=1$, we can write
\begin{align}
    \int q(x)f\left(\frac{p(x)}{q(x)}\right)d\mu(x) &= \int q(x)\left(f\left(\frac{p(x)}{q(x)}\right)-\nu\right)d\mu(x) + \nu\\
    & \geq \int \left[ q(x) (\zeta(x) - \nu ) + \xi(x) \right] d\mu(x) + \nu.
\end{align}
Now, we define the following sets which form a partition of $\Xcal$:
\begin{align}
    L &= \left\{x \in \Xcal : \frac{1}{r_P}p(x) < b \right\},\\
    M &= \left\{x \in \Xcal : b \leq \frac{1}{r_P}p(x) \leq be^\epsilon \right\},\\
    U &= \left\{x \in \Xcal : \frac{1}{r_P}p(x) > be^\epsilon \right\}.
\end{align}
For each case of $x \in L,M,U$, we have $q^*(x)=b,\frac{1}{r_P}p(x),be^\epsilon$, respectively. We observe the following:
\begin{itemize}
    \item If $x \in M$, then $\zeta(x)=\nu$ clearly.
    \item If $x \in U$, then $q^*(x)/p(x) <1/r_P$. Since $(f^*)_+'$ is monotone increasing, we have $\zeta(x) \leq \nu$.
    \item If $x \in L$ and $p(x) \neq 0$, then $q^*(x)/p(x) > 1/r_P$. Again, since $(f^*)_+'$ is monotone increasing, we have $\zeta(x) \geq \nu$.
    \item Finally, if $p(x)=0$ (which implies $x \in L$ also), then from the definition $f^\star(t)=tf(1/t)$, we have $(f^\star)_+'(t) = f(1/t)-\frac{1}{t}f_-'(1/t)$. From \eqref{eq:firstOrderLB} with $y=0$, we have $\nu = (f^\star)_+'(1/r_P) = f(r_P)-r_P f_-'(r_P) \leq f(0) = \zeta(x)$.

    Hence, $\zeta(x) \geq \nu$ for every $x \in L$.

\end{itemize}

Therefore, since $b \leq q(x) \leq be^\epsilon$, we can write 
    \begin{align}
        \int q(x)f\left(\frac{p(x)}{q(x)}\right)d\mu(x) 
        & \geq \int \left[ q(x) (\zeta(x) - \nu ) + \xi(x) \right] d\mu(x) + \nu\\
        & \geq [b(\zeta(x) - \nu )\mathbbm{1}_{L}(x) + be^\epsilon (\zeta(x) - \nu )\mathbbm{1}_{U}(x) + \xi(x)]d\mu(x) + \nu.
    \end{align}
The last expression does not depend on the choice of $q(x)$. Also, we observe that all inequalities become equality if $q(x)=q^*(x)$. This completes the proof for the optimality of $q^*$, and hence completes the proof of the achievability part. $\qed$

\subsection{Proof of converse part (Proposition \ref{prop:converse})}\label{subsec:ConverseProof}
Let $\Qbf \in \mathcal{Q}_{\Xcal,\PtildCal, \epsilon}$ be given. Let $\Acal=\bracketSet{A \subset \Xcal : \mu(A) = \alpha}$. For each $A \in \Acal$, let 
$p_A(x) = \begin{cases}
    c_2 &\text{ if }x \in A\\
    c_1 &\text{ if }x \in \Xcal \backslash A
\end{cases}$.
Since $c_2 \alpha + c_1 (1-\alpha)=1$, we have $\int p_A (x) d\mu (x) = 1$. Hence for each $A \in \Acal$, we can define a probability measure $P_A \in \PtildCal$ by $\frac{dP_A}{d\mu} = p_A$. Also, note that $P_A(A) = c_2 \alpha$.

For each $A \in \Acal$, let $\beta_A = \Qbf(A|P_A)$. Then, the push-forward measures of $P_A$ and $\Qbf(P_A)$ by the indicator function $\mathbbm{1}_A$ are Bernoulli distributions with $\mathrm{Pr}(1)=c_2 \alpha$ and $\beta_A$, respectively. By the data processing inequality (Theorem \ref{thm:fDivDPI}), we have
\begin{align}
    \fDiv{P_A}{\Qbf(P_A)} \geq \fDivBin{c_2 \alpha}{\beta_A}. \label{eq:converseExtDPI}
\end{align}

The main lemma to proceed is the following.
\begin{lemma}\label{lem:decompMinProbBound}
Let $\Xcal$ be a sample space. Let $t,u \in \Nbb$, $t>u$. Let $A_1,A_2,\cdots, A_t \subset \Xcal$ be subsets such that for each $x \in \Xcal$, we have $|\bracketSet{i \in [t] : x \in A_i}| \leq u$. Then for any $t$ probability measures $Q_1,\cdots,Q_t \in \Pcal(\Xcal)$ satisfying that $Q_i(A) \leq e^\epsilon Q_j(A)$ for all $i,j \in [t]$ and $A \subset \Xcal$, we have
\begin{equation}
    \min_{i \in [t]} Q_i(A_i) \leq \frac{(u/t)e^\epsilon}{(u/t) e^\epsilon + 1 - (u/t)}.
\end{equation}
\end{lemma}
Now, by the assumption that $\mu$ is $\alpha$-decomposable,
there exist sequences $\bracketSet{t_j}_{j=1}^{\infty}, \bracketSet{u_j}_{j=1}^{\infty} \subset \Nbb$ of positive integers such that $t_j>u_j$, $\alpha \leq u_j/t_j$, $\lim_{j \rightarrow \infty} u_j/t_j = \alpha$,
and for each $j \in \Nbb$, there exist $t_j$ subsets $A_{j,1},A_{j,2},\cdots,A_{j,t_j} \subset \Xcal$ such that $\mu(A_{j,i})=\alpha$ for all $i \in [t_j]$, and for every $x \in \Xcal$,
we have $\left|\bracketSet{i \in [t_j] : x \in A_{j,t_j}}\right| \leq u_j$. 
By applying Lemma \ref{lem:decompMinProbBound} to $Q_i=\Qbf(P_{A_{j,i}})$, we obtain $\min_{i \in [t_j]} \beta_{A_{i,j}} \leq \frac{(u_j/t_j)e^\epsilon}{(u_j/t_j) e^\epsilon + 1 - (u_j/t_j)}$.
This implies that $\inf_{A \in \Acal} \beta_A \leq \frac{(u_j/t_j)e^\epsilon}{(u_j/t_j) e^\epsilon + 1 - (u_j/t_j)}$ for all $j \in \Nbb$,
and by taking the limit $j \rightarrow \infty$, we have 
$\inf_{A \in \Acal} \beta_A \leq \frac{\alpha e^\epsilon}{\alpha e^\epsilon + 1 - \alpha} = b e^\epsilon \alpha$.

Since $be^\epsilon < c_2$, we have $b e^\epsilon \alpha < c_2 \alpha$. By continuity of $D_f^{\mathrm{B}}$ and the fact that $\fDivBin{\lambda_1}{\lambda_2}$ is decreasing in $\lambda_2 \in [0,\lambda_1]$, we have
\begin{align}
    \sup_{A \in \Acal} \fDiv{P_A}{\Qbf(P_A)} \geq \sup_{A \in \Acal} \fDivBin{c_2 \alpha}{\beta_A} \geq \fDivBin{c_2 \alpha}{b e^\epsilon \alpha}.
\end{align}
It follows that
\begin{align}
    R_f(\Qbf) = \sup_{P \in \PtildCal} \fDiv{P}{\Qbf(P)} \geq \sup_{A \in \Acal} \fDiv{P_A}{\Qbf(P_A)} \geq \fDivBin{c_2 \alpha}{b e^\epsilon \alpha}.
\end{align}
Furthermore, we have
\begin{align}
    \fDivBin{c_2 \alpha}{b e^\epsilon \alpha} & = b e^\epsilon \alpha f\left(\frac{c_2 \alpha}{b e^\epsilon \alpha} \right) + (1-b\alpha e^\epsilon) f\left(\frac{1-c_2 \alpha}{1-b e^\epsilon \alpha} \right)\\
    &= b e^\epsilon \alpha f\left(\frac{c_2 \alpha}{b e^\epsilon \alpha} \right) + b(1-\alpha) f\left(\frac{c_1 (1-\alpha)}{b(1-\alpha)} \right)\\
    &= b e^\epsilon \alpha f(r_2) + b(1-\alpha) f(r_1),
\end{align}
and we can derive $be^\epsilon \alpha = \frac{1-r_1}{r_2 - r_1}$ and $b(1-\alpha)=\frac{r_2 - 1}{r_2-r_1}$, as follows. From \eqref{ineq:c1AndbComp}, \eqref{ineq:c2AndbComp} and the definition of $r_1,r_2$, we have
\begin{align}
    1-r_1 &= (1-c_1) \frac{c_2 - c_1 e^\epsilon}{c_2 - c_1},\\
    r_2 - 1 &= \frac{c_2 - 1}{e^\epsilon} \frac{c_2 - e^\epsilon c_1}{c_2 - c_1}.
\end{align}
From this, we have
\begin{align}
    \frac{1-r_1}{r_2 - r_1} &= \frac{1-r_1}{(r_2-1)+(1-r_1)}\\
    &= \frac{(1-c_1)e^\epsilon}{(1-c_1)e^\epsilon + (c_2-1)}\\
    &= \frac{(1-c_1)e^\epsilon}{(e^\epsilon-1)(1-c_1) + c_2-c_1}\\
    &= b \times e^\epsilon \frac{1-c_1}{c_2 - c_1}\\
    &= b  e^\epsilon \alpha.
\end{align}
Also, from above and $1= be^\epsilon \alpha + b(1-\alpha)$, we have
\begin{align}
    \frac{r_2 - 1}{r_2-r_1} = 1 - \frac{1-r_1}{r_2 - r_1} = 1 - be^\epsilon \alpha = b(1-\alpha).
\end{align}
This ends the proof of the converse part. $\qed$

\subsection{Deduction to main theorems} \label{subsec:deduceMainThm}
In this subsection, we show that Theorem \ref{thm:mainThmGen} contains main theorems, Theorems \ref{thm:optMechFinite} and \ref{thm:optMechCont}, as special cases.
\subsubsection{Deduction to Theorem \ref{thm:optMechFinite}} \label{subsubsec:deduceMainThmFinCase}
Recall that in this setup, $\Xcal=[k]$, $\PtildCal=\Pcal([k])=\PtildCal_{0,k,\mu_k}$, where $\mu_k$ is the uniform distribution on $[k]$. That is, $(c_1,c_2)=(0,k)$. The values of the constants $\alpha, b, r_1, r_2$ are
\begin{align}
    \alpha &= 1/k, \\
    b &= \frac{k}{e^\epsilon + k - 1},\\
    r_1 &= 0,\\
    r_2 &= \frac{e^\epsilon + k - 1}{e^\epsilon}.
\end{align}
We can easily observe that $\mu_k$ is $(\alpha, k, 1)$-decomposable, because the sets $\{i\}$, $i \in [k]$, are disjoint and $\mu_k(\{i\}) = \frac{1}{k}$. It follows that $\mu_k$ is $\alpha$-decomposable. Hence, we can apply Theorem \ref{thm:mainThmGen}. By a direct calculation of $\mathcal{R}(\Xcal,\PtildCal, \epsilon, f) = \frac{1-r_1}{r_2-r_1} f(r_2) + \frac{r_2 - 1}{r_2 - r_1} f(r_1)$, we can derive the formula of $\mathcal{R}([k],\Pcal([k]), \epsilon, f)$ presented in Theorem \ref{thm:optMechFinite}. It remains to show that the mechanism $\Qbf^*_{k,\epsilon}$ presented in Theorem \ref{thm:optMechFinite} is the same as $\Qbf^*_{0,k,\mu_k,\epsilon}$, and show the claim about the range of $r_P$. Recall that
\begin{equation}
    \Qbf^*_{k,\epsilon}(x|P) = \max\left(\frac{1}{r_P} P(x), \frac{1}{e^\epsilon+k-1}\right) \quad \forall x \in [k], P \in \Pcal([k]),
\end{equation}
and we claim that $r_P$ can be chosen such that $1 \leq r_P \leq (e^\epsilon+k-1)/e^\epsilon$. 

First, as explained in Section \ref{subsec:optMechFinSpace}, we can alternatively write
\begin{equation}
    \Qbf^*_{k,\epsilon}(x|P) = \clip \left(\frac{1}{r_P} P(x); \frac{1}{e^\epsilon+k-1}, \frac{e^\epsilon}{e^\epsilon+k-1}\right). \label{eq:optMechFinSpaceAltForm}
\end{equation}
Since $P(x)= \frac{1}{k} \frac{dP}{d\mu_k}(x)$ for each $x \in [k]$ and $P \in \Pcal([k])$, \eqref{eq:optMechFinSpaceAltForm} can be written as
\begin{align}
    \frac{1}{k}\frac{d\Qbf^*_{k,\epsilon}(P)}{d\mu_k} (x) &= \clip \left(\frac{1}{r_P} \frac{1}{k}\frac{dP}{d\mu_k}(x); \frac{1}{e^\epsilon+k-1}, \frac{e^\epsilon}{e^\epsilon+k-1}\right)\\
    & = \frac{1}{k} \clip \left(\frac{1}{r_P} \frac{dP}{d\mu_k}(x); \frac{k}{e^\epsilon+k-1}, \frac{ke^\epsilon}{e^\epsilon+k-1}\right),
\end{align}
hence,
\begin{align}
    \frac{d\Qbf^*_{k,\epsilon}(P)}{d\mu_k} (x) &= \clip \left(\frac{1}{r_P} \frac{dP}{d\mu_k}(x); \frac{k}{e^\epsilon+k-1}, \frac{ke^\epsilon}{e^\epsilon+k-1}\right)\\
    &= \clip \left(\frac{1}{r_P} \frac{dP}{d\mu_k}(x); b, be^\epsilon\right).
\end{align}
Hence, $\Qbf^*_{k,\epsilon} = \Qbf^*_{0,k,\mu_k,\epsilon}$.

Second, to prove the claim about the range of $r_P$, let us fix $P \in \Pcal([k])$. We need to show that
\begin{align}
    \sum_{x=1}^k \max\left(P(x), \frac{1}{e^\epsilon+k-1}\right) \geq 1 \geq \sum_{x=1}^k\max\left(\frac{e^\epsilon}{e^\epsilon+k-1} P(x), \frac{1}{e^\epsilon+k-1}\right).
\end{align}
The left inequality can be easily derived by
\begin{align}
    \sum_{x=1}^k \max\left(P(x), \frac{1}{e^\epsilon+k-1}\right) \geq \sum_{x=1}^k P(x)=1.
\end{align}
For the right inequality, we recall that $b = \frac{k}{e^\epsilon + k - 1}$. Since $P(x) \leq 1$, we have $\frac{e^\epsilon}{e^\epsilon+k-1} P(x) \leq \frac{e^\epsilon}{e^\epsilon+k-1}$. Hence, again by using $P(x)= \frac{1}{k} \frac{dP}{d\mu_k}(x)$, we have
\begin{align}
     &\max\left(\frac{e^\epsilon}{e^\epsilon+k-1} P(x), \frac{1}{e^\epsilon+k-1}\right) \\
    =& \clip \left(\frac{e^\epsilon}{e^\epsilon+k-1} P(x); \frac{1}{e^\epsilon+k-1}, \frac{e^\epsilon}{e^\epsilon+k-1}\right)\\
     =& \clip \left(\frac{e^\epsilon}{e^\epsilon+k-1} \frac{1}{k}\frac{dP}{d\mu_k}(x); \frac{1}{e^\epsilon+k-1}, \frac{e^\epsilon}{e^\epsilon+k-1}\right)\\
     &= \frac{1}{k} \clip \left(\frac{1}{r_2} \frac{dP}{d\mu_k}(x); b, be^\epsilon\right),
\end{align}
and thus
\begin{align}
   & \sum_{x=1}^{k} \max\left(\frac{e^\epsilon}{e^\epsilon+k-1} P(x), \frac{1}{e^\epsilon+k-1}\right)\\
   & = \sum_{x=1}^{k} \frac{1}{k} \clip \left(\frac{1}{r_2} \frac{dP}{d\mu_k}(x); b, be^\epsilon\right)\\
   & = \int \clip \left(\frac{1}{r_2} \frac{dP}{d\mu_k}(x); b, be^\epsilon\right) d\mu_k(x).
\end{align}
The desired inequality follows from Proposition \ref{prop:normFactorRange}.

\subsubsection{Deduction to Theorem \ref{thm:optMechCont}}
Recall that in this setup, $\Xcal=\Rbb^n$, $\PtildCal=\PtildCal_{c_1,c_2,h} = \PtildCal_{c_1,c_2,\mu}$, where $\mu \ll m$ and $d\mu/dm=h$. Note that the normalization condition \eqref{eq:normalizationCond} about $(c_1,c_2,h,\epsilon)$ implies the normalization condition \eqref{eq:normalizationCondGeneral} about $(c_1,c_2,\mu,\epsilon)$.
It can be directly observed that $\Qbf^*_{c_1,c_2,h,\epsilon} = \Qbf^*_{c_1,c_2,\mu,\epsilon}$, because for each $P \in \PtildCal$ with corresponding pdf $p(x)$, the chain rule of the Radon-Nikodym derivative shows that $p(x)=\frac{dP}{dm} = \frac{dP}{d\mu}(x)\frac{d\mu}{dm}(x) = \frac{dP}{d\mu}(x) h(x)$. Hence, once we show that $\mu$ is $\alpha$-decomposable, Theorem \ref{thm:mainThmGen} and Proposition \ref{prop:normFactorRange} directly contain Theorem \ref{thm:optMechCont} as a special case. Thus, it remains to show $\mu$ is $\alpha$-decomposable.

In fact, we prove a stronger statement that: $\mu$ is $(\alpha,t,u)$-decomposable for any $t,u \in \Nbb$ such that $t>u$ and $\alpha \leq u/t$. Then, since the set of rational numbers is dense in $\Rbb$, this also implies that $\mu$ is $\alpha$-decomposable.

To prove this, let us first introduce the following lemma.
\begin{lemma}\label{lem:multiDecompFromOneDecomp}
Let $\alpha \in (0,1)$ and $t,u \in \Nbb$, $t>u$, $\alpha \leq u/t$. If $\mu \in \Pcal(\Xcal)$ is $(\alpha/u, t, 1)$-decomposable, then $\mu$ is also $(\alpha,t,u)$-decomposable.
\end{lemma}
\begin{proof}
In this proof, assume that the sum and subtraction operations performed in subscripts are modulo $t$ operations, with the identification that $0=t$.

By $(\alpha/u, t, 1)$-decomposability, there are $t$ disjoint subsets $B_1,B_2,\cdots,B_{t}$ such that $\mu(B_i)=\alpha/u$ for each $i \in [t]$. 
Using this, for each $i \in [t]$, define $A_i$ as $A_i = \cup_{j=0}^{u-1} B_{i+j}$. As $B_i$'s are disjoint, we have $\mu(A_i) = \sum_{j=0}^{u-1} \mu (B_{i+j}) = u \times (\alpha/u) = \alpha$ for all $i \in [t]$. Also, for each $x \in B_i$, $x$ is contained in exactly $u$ sets among $A_1,\cdots,A_t$, which are $A_{i},A_{i- 1},\cdots,A_{i-u+1}$. Furthermore, if $x \notin B_i$ for all $i \in [t]$, then $x$ is contained in none of $A_i$. Hence $|\bracketSet{i \in [t] : x \in A_i}| \leq u$ for all $x \in \Xcal$. Thus  $\mu$ is $(\alpha,t,u)$-decomposable.
\end{proof}
By this lemma, it suffices to show that for any $t \in \Nbb$ such that $t \geq 2$ and $\alpha \leq 1/t$, $\mu$ is $(\alpha, t, 1)$-decomposable. As $\mu \ll m$, the map $s \in \Rbb \mapsto \mu((-\infty,s] \times \Rbb^{n-1})$ is continuous, and as $s \rightarrow - \infty$ and $\infty$, we have $\mu((-\infty,s] \rightarrow 0$ and $1$, respectively. Hence by the intermediate value theorem, for each $i \in [t]$, there exists $s_i \in \Rbb$ such that $\mu((-\infty,s] \times \Rbb^{n-1}) = \alpha i$. Then, setting $A_1 = (-\infty, s_1] \times \Rbb^{n-1}$ and $A_i = (s_{i-1},s_i] \times \Rbb^{n-1}$ for $i \geq 2$ gives the desired $A_i$'s in the definition of decomposability.

In conclusion, $\mu$ is $\alpha$-decomposable, and hence Theorem \ref{thm:optMechCont} can be deduced from Theorem \ref{thm:mainThmGen}.

\subsection{Proof of Proposition \ref{prop:normFactorRange}}\label{subsec:proofPropNormFactorRange}
Let $P \in \PtildCal$ be given. Let $p=\frac{dP}{d\mu}$, and again assume that $c_1 \leq p(x) \leq c_2$ for \emph{all} $x \in \Xcal$. Similar to the proof of the achievability part, for each $r>0$, let us define the following sets which form a partition of $\Xcal$:
\begin{align}
    L_r &= \left\{x \in \Xcal : \frac{1}{r}p(x) < b\right\},\\
    M_r &= \left\{x \in \Xcal : b \leq \frac{1}{r}p(x) \leq be^\epsilon\right\},\\
    U_r &= \left\{x \in \Xcal : \frac{1}{r}p(x) > be^\epsilon\right\}.
\end{align}
For $r=0$, we let $L_0 = \bracketSet{x \in \Xcal: p(x)=0}$, $U_0 = \bracketSet{x \in \Xcal: p(x)>0}$, $M_0 = \emptyset$. Also, let $q_r(x)=\clip\left(\frac{1}{r} p(x); b,  be^\epsilon \right)$. Then, $q_r(x) = b, \frac{1}{r}p(x), be^\epsilon$ for $x \in L_r, M_r, U_r$, respectively.

First, we show that $\int q_{r_1} (x) d\mu(x) \geq 1$. We divide the case of $c_1=0$ and $c_1>0$.

    Suppose first that $c_1=0$. Then $r_1=0, \alpha = \frac{1}{c_2}$, and $b=\frac{c_2}{e^\epsilon + c_2 - 1}$. Observe that
    \begin{align}
        1 = \int p(x) d\mu(x) = \int_{U_0} p(x) d\mu(x) \leq c_2 \mu (U_0),
    \end{align}
    hence $\mu(U_0) \geq 1/c_2$. It follows that
    \begin{align}
        \int q_{r_1} (x) d\mu(x) &= b\mu(L_0) + be^\epsilon \mu(U_0)\\
        &= b(1-\mu(U_0)) + be^\epsilon\mu(U_0)\\
        &= b{(e^\epsilon - 1)}\mu(U_0) + b\\
        & \geq \frac{b(e^\epsilon - 1)}{c_2} + b\\
        &= b \times \frac{e^\epsilon + c_2 - 1}{c_2} = 1.
    \end{align}

    Next, suppose that $c_1 > 0$. Then $r_1 > 0$. From $p(x) \geq c_1$, we have $\frac{1}{r_1}p(x) \geq \frac{c_1}{r_1}=b$. Hence $L_{r_1} = \emptyset$. Thus
    \begin{align}
        \int q_{ r_1}(x)d\mu(x) &= \frac{1}{r_1}\int_{M_{r_1}}p(x)d\mu(x) + be^\epsilon \mu(U_{r_1})\\
        &= \frac{1}{r_1} \left(\int_{M_{r_1}}p(x)d\mu(x) + c_1 e^\epsilon  \mu(U_{r_1}) \right).
    \end{align}
    Let $S_1=\int_{M_{r_1}}p(x)d\mu(x)$ and $T_1=\mu(U_{r_1})$, so that
    \begin{equation}
        \int q_{ r_1}(x)d\mu(x) = \frac{1}{r_1} (S_1 + c_1 e^\epsilon T_1).
    \end{equation}
    As $c_1 \leq p(x) \leq c_2$, we have
    \begin{align}
        S_1=\int_{M_{r_1}} p(x) d\mu(x) \geq c_1 \mu(M_{r_1}) = c_1 (1-\mu(U_{r_1})) = c_1 (1-T_1), \label{ineq:S1Bdd1}
    \end{align}
    and
    \begin{align}
      1- S_1 =  1 - \int_{M_{r_1}}p(x) d\mu(x)=  \int_{U_{r_1}} p(x) d\mu(x) \leq c_2 \mu(U_{r_1}) = c_2 T_1. \label{ineq:S1Bdd2}
    \end{align}
    From these, we can get
    \begin{align}
        S_1 + c_1 T_1 &\geq c_1, \label{eq:rLowBddLp1}\\
        S_1 + c_2 T_1 &\geq 1.\label{eq:rLowBddLp2}
    \end{align}
    As $c_1 < c_1 e^\epsilon < c_2$, we can express $c_1 e^\epsilon$ as a convex combination of $c_1$ and $c_2$, as
    \begin{equation}
        c_1 e^\epsilon = \frac{c_2 - c_1 e^\epsilon}{c_2 - c_1}c_1 + \frac{c_1 e^\epsilon - c_1}{c_2 - c_1} c_2.
    \end{equation}
    Hence, by taking $\frac{c_2 - c_1 e^\epsilon}{c_2 - c_1} [\text{equation }\eqref{eq:rLowBddLp1}] + \frac{c_1 e^\epsilon - c_1}{c_2 - c_1} [\text{equation }\eqref{eq:rLowBddLp2}]$, we get
    \begin{align}
        S + c_1 e^\epsilon T &\geq \frac{c_2 - c_1 e^\epsilon}{c_2 - c_1} c_1 + \frac{c_1 e^\epsilon - c_1}{c_2 - c_1}\\
        & = \frac{c_2 - c_1 e^\epsilon + e^\epsilon - 1}{c_2 - c_1}c_1\\
        &= \frac{(e^\epsilon-1)(1-c_1)+c_2-c_1}{c_2 - c_1} c_1\\
        &= r_1.
    \end{align}
    Thus we have $\int q_{ r_1}(x)d\mu(x) \geq 1$.

    Similarly, we show that $\int q_{r_2} (x) d\mu(x) \leq 1$.  from $p(x) \leq c_2$, we have $\frac{1}{r_2}p(x) \leq \frac{c_2}{r_2}=be^\epsilon$. Hence $U_{r_2} = \emptyset$. Thus
    \begin{align}
        \int q_{ r_2}(x)d\mu(x) &=  b {\mu}(L_{r_2}) + \frac{1}{r_2}\int_{M_{r_2}}p(x)d\mu(x)\\
        &= \frac{1}{r_2} \left(c_2 e^{-\epsilon}{\mu}(L_{r_2}) +  \int_{M_{r_2}}p(x)d\mu(x)\right).
    \end{align}
    Similar as above, let ${S_2}={\mu}(L_{r_2})$ and ${T_2}=\int_{M_{r_2}}p(x)d\mu(x)$, so that
    \begin{align}
        \int q_{ r_2}(x)d\mu(x) = \frac{1}{r_2}(c_2 e^{-\epsilon}{S_2} + {T_2}),
    \end{align}
    and, we have
    \begin{align}
        {T_2} = \int_{M_{r_2}}p(x)d\mu(x) \leq c_2 {\mu}(M_{r_2}) = c_2 (1-{\mu}(L_{r_2})) = c_2 (1-{S_2})
    \end{align}
    and
    \begin{align}
        1-{T_2}=1-\int_{M_{r_2}}p(x)d\mu(x) = \int_{L_{r_2}} p(x)d\mu(x) \geq c_1 {\mu}(L_{r_2}) = c_1 {S_2}.
    \end{align}
    hence
    \begin{align}
        c_2 {S_2} + {T_2} &\leq c_2, \\
        c_1 {S_2} + {T_2} &\leq 1.
    \end{align}
    As $c_1 \leq c_2 e^{-\epsilon} \leq c_2$, we have
    \begin{equation}
        c_2 e^{-\epsilon} = \frac{c_2 - c_2 e^{-\epsilon}}{c_2 - c_1} c_1 + \frac{c_2 e^{-\epsilon}-c_1}{c_2 - c_1}c_2
    \end{equation}
    and by the same reason, we have
    \begin{align}
        c_2 e^{-\epsilon}{S_2} + {T_2} &\leq \frac{c_2 - c_2 e^{-\epsilon}}{c_2 - c_1} + \frac{c_2 e^{-\epsilon}-c_1}{c_2 - c_1}c_2 \\
        &= \frac{1-e^{-\epsilon} + c_2 e^{-\epsilon}-c_1}{c_2 - c_1} c_2\\
        &= \frac{(e^\epsilon-1)(1-c_1)+c_2-c_1}{c_2 - c_1} c_2 e^{-\epsilon}\\
        &= r_2.
    \end{align}
    Thus we have $\int q_{ r_2}(x)d\mu(x) \leq 1$. 
    
    Finally, we show that $\int q_{r_1} (x) d\mu(x) = 1$ implies $\int q_{r} (x) d\mu(x) = 1$ for all $r \in [r_1,r_2]$. Let us assume $\int q_{r_1} (x) d\mu(x) = 1$. We first claim that: $p(x)=c_2$ for $\mu$-a.e. $x \in U_{r_1}$, and $p(x)=c_1$ for $\mu$-a.e. $x \in \Xcal \backslash U_{r_1}$. Again, we divide the case of $c_1=0$ and $c_1>0$.

    Suppose first that $c_1=0$. By tracking the proof of $\int q_{r_1} (x) d\mu(x) \geq 1$, We can observe that the equality $\int q_{r_1} (x) d\mu(x) = 1$ holds if and only if $\mu(U_0)=1/c_2$, if and only if $p(x)=c_2$ for $\mu$-a.e. $x \in U_0$. By definition of $U_0$, we have $p(x)=0=c_1$ for all $x \in \Xcal \backslash U_0$. Hence we get the claim.

    Next, suppose that $c_1>0$. Again, by tracking the proof of $\int q_{r_1} (x) d\mu(x) \geq 1$, We can observe that the equality $\int q_{r_1} (x) d\mu(x) = 1$ is equivalent to any of the following statements:
    \begin{itemize}
        \item Both $S_1 + c_1 T_1 = c_1$ and $S_1 + c_2 T_1 = 1$ holds.
        \item Both $\int_{M_{r_1}} p(x) d\mu(x) = c_1 \mu(M_{r_1})$ and $\int_{U_{r_1}} p(x)d\mu(x) = c_2 \mu(U_{r_1})$ holds.
        \item $p(x)=c_1$ for $\mu$-a.e. $x \in M_{r_1}$ and $p(x)=c_2$ for $\mu$-a.e. $x \in U_{r_1}$.
    \end{itemize}
    Since $L_{r_1}=\emptyset$, we also get the claim.

    WLOG, assume that $p(x)=c_2$ for \emph{all} $x \in U_{r_1}$, and $p(x)=c_1$ for \emph{all} $x \in \Xcal \backslash U_{r_1}$. Then, for every $r \in (r_1, r_2]$, we have the following:
    \begin{itemize}
        \item For each $x \in U_{r_1}$, we have $\frac{1}{r}p(x) \geq \frac{c_2}{r_2}=be^\epsilon$, hence $q_r(x)=be^\epsilon$.
        \item For each $x \in \Xcal \backslash U_{r_1}$,
        \begin{itemize}
            \item If $c_1=0$, then $p(x)=0$, $q_r(x)=b$, and
            \item If $c_1>0$, then $r_1>0$, $\frac{1}{r}p(x) < \frac{c_1}{r_1}=b$, hence again $q_r(x)=b$.
        \end{itemize}
    \end{itemize}
    Also, we have $1=\int p(x) d\mu(x) = c_2 \mu(U_{r_1})+c_1(1-\mu(U_{r_1}))$, hence $\mu(U_{r_1})=\frac{1-c_1}{c_2-c_1}=\alpha$. It follows that for every $r \in (r_1,r_2]$, we have $\int q_r(x) d\mu(x)=be^\epsilon \mu(U_{r_1})+b(1-\mu(U_{r_1})) = be^\epsilon \alpha + b(1-\alpha)=1$. This concludes the proof.  $\qed$

\subsection{Proof of Lemma \ref{lem:densityRatioBddToDivBdd}} \label{subsec:proofLemDensityRatioBddToDivBdd}
    Although the proof can be found in \cite{Rukhin97-fDivBddForLRBdd}, we present the proof here for the completeness.
    
    If $r_1=0$ and $f(0)=\infty$, then the RHS of the inequality we want to show is $\infty$, thus it becomes trivial. Hence, we may assume that either $r_1>0$ or $f(0)<\infty$.

    By the assumption, $p(x)/q(x)$ is the convex combination of ${r_1}$ and ${r_2}$ for $\mu$-a.e. $x \in \mathcal{X}$, as follows.
    \begin{equation}
        p(x)/q(x) = \frac{(p(x)/q(x))-{r_1}}{{r_2}-{r_1}}{r_2} + \frac{{r_2}-(p(x)/q(x))}{{r_2}-{r_1}}{r_1}.
    \end{equation}
    Hence, by the convexity of $f$ and $\int p(x)d\mu(x)=\int q(x)d\mu(x)=1$, we have
    \begin{align}
        \fDiv{P}{Q} &= \int q(x) f(p(x)/q(x)) d\mu(x)\\
        & \leq \int q(x) \left(\frac{(p(x)/q(x))-{r_1}}{{r_2}-{r_1}} f({r_2}) + \frac{{r_2}-(p(x)/q(x))}{{r_2}-{r_1}} f({r_1})  \right) d\mu(x)\\
        &= \int \left({\frac{p(x)-r_1 q(x)}{{r_2}-{r_1}}} f({r_2}) + \frac{r_2 q(x)-p(x)}{{r_2}-{r_1}} f({r_1})\right)d\mu(x)\\
        &= \frac{1-{r_1}}{{r_2}-{r_1}}f({r_2}) + \frac{{r_2}-1}{{r_2}-{r_1}}f({r_1}).
    \end{align}
$\qed$

\subsection{Proof of Lemma \ref{lem:decompMinProbBound}}\label{subsec:proofLemDecompMinProbBound}
First, we claim that for each $i \in [t-u]$, we can construct a partition $\bracketSet{B_{i,1},B_{i,2},\cdots,B_{i,u}}$ of $A_{u+i}$ into $u$ (measurable) sets, such that for each $j \in [u]$, the sets $A_j, B_{1,j}, B_{2,j}, \cdots, B_{t-u, j}$ are disjoint. The construction is in the inductive way as follows: Given $i \in [t-u]$, suppose that we have constructed such partitions of $A_{u+1},\cdots,A_{u+i-1}$ such that for each $j \in [u]$, $A_j, B_{1,j}, B_{2,j}, \cdots, B_{i-1, j}$ are disjoint. Let $C_{i,j} = A_j \cup \bracket{\cup_{k=1}^{i-1} B_{k,j}}$. Then for each $x \in A_{u+i}$, at least one of $C_{i,1},C_{i,2},\cdots,C_{i,u}$ does not contain $x$, because
\begin{align}
    \sum_{j=1}^u \mathbbm{1}_{\Xcal \backslash C_{i,j}}(x) &= \sum_{j=1}^u \bracket{1-\mathbbm{1}_{C_{i,j}}(x)} = u - \sum_{j=1}^u {\mathbbm{1}_{C_{i,j}}(x)}
\end{align}
\begin{align}
    = u - \sum_{j=1}^u \bracket{\mathbbm{1}_{A_j}(x) + \sum_{k=1}^{i-1} \mathbbm{1}_{B_{k,j}}(x)}
    &= u - \sum_{j=1}^u \mathbbm{1}_{A_j}(x) - \sum_{j=1}^u \sum_{k=1}^{i-1} \mathbbm{1}_{B_{k,j}}(x)\\
    = u - \sum_{j=1}^u \mathbbm{1}_{A_j}(x) -  \sum_{k=1}^{i-1}\sum_{j=1}^u \mathbbm{1}_{B_{k,j}}(x)
    &= u - \sum_{j=1}^u \mathbbm{1}_{A_j}(x) -  \sum_{k=1}^{i-1} \mathbbm{1}_{A_{u+k}}(x)\\
    & \geq u - (u-1) = 1,
\end{align}
where the last line is from that since $x \in A_{u+i}$, at most $u-1$ sets among $A_1,A_2,\cdots,A_{u+i-1}$ can contain $x$. Hence, setting
\begin{equation}
    B_{i,j} = \bracketSet{x \in A_{u+i} : j = \min \bracket{\tilde{j} \in [u]: x \notin C_{i,\tilde{j}}}}
\end{equation}
gives the partition $\bracketSet{B_{i,1},B_{i,2},\cdots,B_{i,u}}$ of $A_{u+i}$, such that for each $j \in [u]$, $C_{i,j}$ and $B_{i,j}$ are disjoint. This implies that $A_j, B_{1,j}, B_{2,j}, \cdots, B_{i, j}$ are disjoint.

Now, let $\ell = \min_{i \in [t]} Q_i(A_i)$. Using these partitions, we can now show that
\begin{align}
    u &= \sum_{j=1}^{u} Q_j (\Xcal) \geq \sum_{j=1}^{u} Q_j \bracket{A_j \cup \bracket{\bigcup_{k=1}^{t-u} B_{k,j}}}\\
    &= \sum_{j=1}^{u} Q_j(A_j) + \sum_{j=1}^{u} \sum_{k=1}^{t-u} Q_j(B_{k,j}) = \sum_{j=1}^{u} Q_j(A_j) + \sum_{k=1}^{t-u}\sum_{j=1}^{u} Q_j(B_{k,j})\\
    &= \sum_{j=1}^{u} Q_j(A_j) + \sum_{k=1}^{t-u} Q_j(A_{u+k})\\
    &\geq \sum_{j=1}^{u} Q_j(A_j) + \sum_{k=1}^{t-u} e^{-\epsilon} Q_{u+k}(A_{u+k})\\
    &\geq \ell\bracket{u + e^{-\epsilon}(t-u)}.
\end{align}
Hence
\begin{align}
    \ell \leq \frac{u}{u + e^{-\epsilon}(t-u)} = \frac{(u/t)e^\epsilon}{(u/t) e^\epsilon + 1 - (u/t)}.
\end{align}
$\qed$

\section{Proofs of Remaining Propositions}\label{supp:remainPropStateAndProof}
We present the proofs of remaining propositions in the paper, Propositions \ref{prop:WorstFDivAlwaysMax} and \ref{prop:mechContCoeffBdd}.

\subsection{Proof of Proposition \ref{prop:WorstFDivAlwaysMax}}
By the assumption, there are subsets $\bracketSet{A_i}_{i=1}^{\infty}$ of $\Xcal$ which are pairwise disjoint and $P_i(A_i)=1$ for all $i \in \Nbb$. Let us pick $P_0 \in \PtildCal$, and let $Q_0 = \Qbf(P_0)$. Since $A_i$'s are disjoint, we have $\sum_{i=1}^{\infty}Q_0(A_i) = Q_0 \bracket{\bigcup_{i=1}^{\infty}A_i} \leq 1 < \infty$. Hence, we have $\lim_{i \rightarrow \infty}Q_0(A_i)=0$. Also, by definition of $\epsilon$-LDP, for any $P \in \PtildCal$ and $i \in \mathbb{N}$, we have $\mathbf{Q}(P)(A_i) \leq e^\epsilon Q_0 (A_i)$. Especially, this implies $\mathbf{Q}(P_i)(A_i) \leq e^\epsilon Q_0 (A_i)$, and thus $\lim_{i \rightarrow \infty}\mathbf{Q}(P_i)(A_i) = 0$.

Now, similar to the converse proof in Section \ref{subsec:ConverseProof}, let $\beta_i = \mathbf{Q}(P_i)(A_i)$. Then, the push-forward measures of $P_i$ and $\Qbf(P_i)$ by the indicator function $\mathbbm{1}_A$ are Bernoulli distributions with $\mathrm{Pr}(1)=1$ and $\beta_i$, respectively. By the data processing inequality (Theorem \ref{thm:fDivDPI}), we have
\begin{align}
    \fDiv{P_i}{\mathbf{Q}(P_i)} & \geq \fDivBin{1}{\beta_i}.
\end{align}
Since $\lim_{i \rightarrow \infty} \beta_i=0$, by continuity of $D_f^{\mathrm{B}}$, we have
\begin{align}
    R_f(\Qbf) \geq \limsup_{i \rightarrow \infty} \fDiv{P_i}{\mathbf{Q}(P_i)} \geq \lim_{i \rightarrow \infty} \fDivBin{1}{\beta_i} = \fDivBin{1}{0} = M_f,
\end{align}
where the last equality is because two Bernoulli distributions with $\mathrm{Pr}(1)=1$ and $\mathrm{Pr}(1)=0$, respectively, are mutually singular.
Hence, we must have $R_f(\Qbf) = M_f$. $\qed$

\subsection{Proof of Proposition \ref{prop:mechContCoeffBdd}}
For generality, we prove that the statement of Proposition \ref{prop:mechContCoeffBdd} holds in the general setup in Definition \ref{def:mechDefGenSetup}. That is, for the setup in Definition \ref{def:mechDefGenSetup}, the mechansim $\Qbf^* = \Qbf^*_{c_1,c_2,\mu,\epsilon}$ satisfies
\begin{equation}
    \tvDiv{\Qbf^*(P)}{\Qbf^*(P')} \leq \frac{2}{\max(r_P, r_{P'})} \tvDiv{P}{P'}
\end{equation}
for all $P,P' \in \PtildCal$. As the mechanisms $\Qbf^*_{k,\epsilon}$ and $\Qbf^*_{c_1,c_2,h,\epsilon}$ in the paper are special cases of $\Qbf^*_{c_1,c_2,\mu,\epsilon}$, a proof in this general setup induces Proposition \ref{prop:mechContCoeffBdd}. 

Now, let us assume the setup in Definition \ref{def:mechDefGenSetup} Let $P,P' \in \PtildCal$ be given. WLOG, assume that $r_P \geq r_{P'}$. Let $p=dP/d\mu$, $p'=dP'/d\mu$, and $q(x)=\clip\left(\frac{1}{r_P}p(x);b,be^\epsilon \right)$, $q'(x)=\clip\left(\frac{1}{r_{P'}}p'(x);b,be^\epsilon \right)$, so that $q=d\Qbf^*(P)/d\mu$ and $q'=d\Qbf^*(P')/d\mu$. For simplicity, we denote $\clip(x):=\clip(x;b,be^\epsilon)$.

We first note the fact that $\clip(x)$ is monotone increasing and 1-Lipschitz in $x$. From this and the equivalent expressions of the total variation distance in Appendix \ref{supp:fDivReview}, we have

\begin{align}
    & \tvDiv{\Qbf^*(P)}{\Qbf^*(P')}\\
    & = \int_{x:q(x) \geq q'(x)} (q(x) - q'(x)) d\mu(x)\\
    & = \int_{x:q(x) \geq q'(x)} \bracket{\clip\bracket{\frac{1}{r_P}p(x)} - \clip\bracket{\frac{1}{r_{P'}}p'(x)}}d\mu(x)\\
    & = \int_{x:q(x) \geq q'(x)} \bracket{\clip\bracket{\frac{1}{r_P}p(x)} - \clip\bracket{\frac{1}{r_P}p'(x)}}d\mu(x) \nonumber\\
    & \quad + \int_{x:q(x) \geq q'(x)} \bracket{\clip\bracket{\frac{1}{r_P}p'(x)} - \clip\bracket{\frac{1}{r_{P'}}p'(x)}}d\mu(x)\\
    & \leq \int_{x:q(x) \geq q'(x)} \left|\frac{1}{r_P}(p(x)-p'(x))\right|d\mu(x) + 0 \label{ineq:clipMonotoneAndLip}\\
    & \leq \frac{1}{r_P} \int_{\Xcal} |p(x)-p'(x)| d\mu(x) = \frac{2}{r_P} \tvDiv{P}{P'}.
\end{align}
This ends the proof. $\qed$

\section{Behaviors of Proposed Mechanisms} \label{supp:mechBehavior}
In this appendix, we present the formal proofs for the behaviors of the proposed mechanisms presented in Sections \ref{subsec:optMechFinSpace} and \ref{subsec:contSpace}.

We first observe that the formula of the optimal worst-case $f$-divergence in general case
\begin{equation}
    \frac{1-r_1}{r_2-r_1} f(r_2) + \frac{r_2 - 1}{r_2 - r_1} f(r_1) \label{eq:optWorstfDivFormulaOnly}
\end{equation}
is the $y$-coordinate value at $x=1$ of the line segment joining $(r_1,f(r_1))$ and $(r_2,f(r_2))$. As $f$ is convex, this formula is increasing in $r_2$ and decreasing in $r_1$, provided that $r_1<1<r_2$.

Now, let us present the proofs.

\begin{itemize}
    \item If $f(0)=\infty$ and $\PtildCal$ contains two mutually singular distributions, then $\Rcal(\Xcal, \PtildCal, \epsilon, f)=\infty$.
    \begin{proof}
        Let $P_1,P_2 \in \PtildCal$ be mutually singular distributions with disjoint supports $A_1,A_2 \subset \Xcal$ respectively (That is, $P_1(A_1)=P_2(A_2)=1$ and $A_1 \cap A_2 = \emptyset$). Suppose that $\Qbf$ is an $\epsilon$-LDP sampling mechanism for $(\Xcal, \PtildCal)$ such that $R_f(\Qbf)<\infty$. Since $f(0)=\infty$, $\fDiv{P}{Q}<\infty$, implies $Q \ll P$. Hence, as $\fDiv{P_i}{ \Qbf(P_i)}<\infty$ and $P_i(A_i^c) = 0$, we have $\Qbf(A_i^c|P_i)=0$ for each $i=1,2$. As $\Qbf$ satisfies $\epsilon$-LDP, we have $\Qbf(A_1^c | P_2) \leq \Qbf(A_1^c|P_1)=0$, $\Qbf(A_1^c | P_2)=0$. Now, since $A_1 \cap A_2 = \emptyset$, we have $A_1^c \cup A_2^c = \Xcal$. But by the union bound, $1=\Qbf(\Xcal|P_2) \leq \Qbf(A_1^c|P_2)+\Qbf(A_2^c|P_2)=0$, which is a contradiction. Hence, $R_f(\Qbf)=\infty$ for every $\epsilon$-LDP sampling mechanism $\Qbf$.
    \end{proof}
\end{itemize}

From now, assume $f(0)<\infty$.

Let us first prove the behaviors for the case of finite $\Xcal$ in Section \ref{subsec:optMechFinSpace}.
\begin{itemize}
    \item $\mathcal{R}([k],\Pcal([k]), \epsilon, f)$ is decreasing in $\epsilon$ and increasing in $k$.
    \begin{proof}
        Recall from Appendix \ref{subsubsec:deduceMainThmFinCase} that the case of $\Xcal=[k]$, $\PtildCal=\Pcal([k])$ can be fit into the general case with
        \begin{align}
            r_1 &= 0,\\
            r_2 &= \frac{e^\epsilon + k - 1}{e^\epsilon}.
        \end{align}
        Here, $r_2$ is decreasing in $\epsilon$ and increasing in $k$. As \eqref{eq:optWorstfDivFormulaOnly} is increasing in $r_2$, we get the desired claim.
    \end{proof}
    
    \item For a fixed $k$, we have $\mathcal{R}([k],\Pcal([k]), \epsilon, f) \rightarrow 0$ as $\epsilon \rightarrow \infty$.
    \begin{proof}
        As $\epsilon \rightarrow \infty$, we have $r_2 \rightarrow 1$. As $f$ is continuous, $f(1)=0$, and $f(0)<\infty$, we obtain from \eqref{eq:optWorstfDivFormulaOnly} that
        \begin{align}
            \mathcal{R}([k],\Pcal([k]), \epsilon, f) \rightarrow \frac{1-0}{1-0}f(1) + \frac{1-1}{1-0}f(0)=0 \label{eq:worstfDivAsEpsInfty}
        \end{align}
        as $\epsilon \rightarrow \infty$.
    \end{proof}

    \item For a fixed $k$, as $\epsilon \rightarrow 0$, we have $\Qbf^*_{k,\epsilon}(x|P) \rightarrow 1/k$ for every $P \in \Pcal([k])$ and $x \in [k]$.
    \begin{proof}
        We know that $\frac{1}{e^\epsilon+k-1} \leq \Qbf^*_{k,\epsilon}(x|P) \leq \frac{e^\epsilon}{e^\epsilon+k-1}$. As $\epsilon \rightarrow 0$, both of $\frac{1}{e^\epsilon+k-1}$ and $\frac{e^\epsilon}{e^\epsilon+k-1}$ converges to $\frac{1}{k}$, hence we get the desired claim.
    \end{proof}
\end{itemize}
Next, let us prove the behaviors for the continuous case in Section \ref{subsec:contSpace}.
\begin{itemize}
    \item If $c_1=0$ and $f(0)=\infty$, then $\mathcal{R}(\Rbb^n,\PtildCal_{c_1, c_2, h}, \epsilon, f)=\infty$.

    \begin{proof}
        Since $r_2>1$ and $r_1=0$, we have $\frac{r_2-1}{r_2-r_1}=\frac{r_2-1}{r_2}>0$. Hence $\frac{r_2-1}{r_2-r_1}f(r_1) = \frac{r_2-1}{r_2-r_1} f(0)=\infty$, which proves $\mathcal{R}(\Rbb^n,\PtildCal_{c_1, c_2, h}, \epsilon, f)=\infty$.
    \end{proof}

    \item For a fixed $(c_1,c_2)$, $\mathcal{R}(\Rbb^n,\PtildCal_{c_1, c_2, h}, \epsilon, f)$ is decreasing in $\epsilon$.
    \begin{proof}
        We can observe that $r_1$ is increasing in $\epsilon$ and $r_2$ is decreasing in $\epsilon$. Since \eqref{eq:optWorstfDivFormulaOnly} is increasing in $r_2$ and decreasing in $r_1$, we get the desired claim.
    \end{proof}
    \item For a fixed $(c_1,c_2)$ with $c_1=0$, as $\epsilon \rightarrow \infty$, we have  $\mathcal{R}(\Rbb^n,\PtildCal_{c_1, c_2, h}, \epsilon, f) \rightarrow 0$.
    \begin{proof}
        We can observe that $r_1=0$ and $r_2 \rightarrow 1$. Hence, by the same argument as \eqref{eq:worstfDivAsEpsInfty}, we have the desired claim.
    \end{proof}
\end{itemize}

\section{Detailed Explanation of Setups in Numerical Results}\label{supp:numResultSetups}
We present details about the setups in producing numerical results in Section \ref{sec:numResult}, and generating Figure \ref{fig:ringGauss} in Section \ref{sec:intro}.

In this appendix, for each $\mu \in \Rbb^n$, $\mathcal{N}_{\mu,\Sigma}[x] = \frac{1}{{(2\pi)}^{n/2} \sqrt{\det(\Sigma)}} \exp\left(-\frac{(x-\mu)^T \Sigma^{-1} (x-\mu)}{2}\right)$ is the pdf of the $n$-dimensional Gaussian distribution with mean $\mu$ and covariance $\Sigma$. Note that we denote $\mathcal{N}(\mu, \Sigma)$ to refer the Gaussian distribution itself with mean $\mu$ and covariance matrix $\Sigma$.

\subsection{Explanation for numerical result for finite data space}\label{supp:numResultSetupsFinSpace}
In this appendix, we show that among the choices of $Q_0$ in the baseline for $\Xcal=[k]$, choosing $Q_0$ to be the uniform distribution minimizes $R_f(\Qbf)$, and present the precise value of $R_f(\Qbf)$ for the baseline with uniform $Q_0$. From now, let us fix $k,\epsilon,f$, and we denote $\Qbf_{Q_0}$ to mean the baseline with the reference distribution $Q_0$. Also, let $\delta_x \in \Pcal([k])$ be the point mass at $x \in [k]$. Recall that
\begin{align}
    \Mcal_{\epsilon, Q_0} = \{Q \in \Pcal(\Xcal): e^{-\epsilon/2} Q_0(A) \leq Q(A) \leq e^{\epsilon/2} Q_0(A), \quad \forall A \subset \Xcal\},
\end{align}
and we set the baseline to satisfy that
\begin{align}
    \fDiv{P}{\Qbf_{Q_0}(P)} = \inf_{Q \in \Mcal_{\epsilon, Q_0}} \fDiv{P}{Q}.
\end{align}
First, if $Q_0(x)=0$ for some $x \in [k]$, then $\Qbf_{Q_0}(x|P) \leq e^{\epsilon/2} Q_0(x)=0$, $\Qbf_{Q_0}(x|P)=0$ for every $P \in \Pcal([k])$. Especially, $\Qbf_{Q_0}(x|\delta_x)=0$, and this implies that $\Qbf_{Q_0}(\delta_x)$ and $\delta_x$ are mutually singular. Hence,
\begin{align}
    R_f(\Qbf_{Q_0}) \geq \fDiv{\delta_x}{\Qbf_{Q_0}(\delta_x)} = M_f,
\end{align}
concluding that $R_f(\Qbf_{Q_0})=M_f$. Hence, to minimize $R_f(\Qbf_{Q_0})$, it suffices to set $Q_0(x)>0$ for all $x \in [k]$. Hence from now, we only consider such $Q_0$.

We note that for every $x \in [k]$ and $P \in \Pcal([k])$, we have we have $\Qbf_{Q_0}(x|P) \leq e^{\epsilon/2} Q_0(x)$, and furthermore
\begin{align}
    \Qbf_{Q_0}(x|P) & = 1-\sum_{y \in [k] \backslash \{x\}}  \Qbf_{Q_0}(y|P) \label{eq:QBoundPrevInit}\\
    & \leq 1 - \sum_{y \in [k] \backslash \{x\}}  e^{-\epsilon/2} Q_0(y)\\
    & = 1 - e^{-\epsilon/2}(1-Q_0(x))\\
    & = e^{-\epsilon/2}Q_0(x) + (1-e^{-\epsilon/2}). \label{eq:QBoundPrevFinal}
\end{align}
Letting 
\begin{align}
    B(t)=\min\{e^{\epsilon/2}t, e^{-\epsilon/2}t + (1-e^{-\epsilon/2})\},
\end{align}
we have
\begin{align}
    \Qbf_{Q_0}(x|P) \leq B(Q_0(x)). \label{eq:PerturbPmfLB}
\end{align}
Note that $B(t)$ is increasing in $t$.
Now, for any $x \in [k]$, we have
\begin{align}
    R_f(\Qbf_{Q_0}) &\geq \fDiv{\delta_x}{\Qbf_{Q_0}(\delta_x)} \label{eq:WorstfDivFiniteLBInit}\\
    & = \Qbf_{Q_0}(x|\delta_x) f\left(\frac{1}{\Qbf_{Q_0}(x|\delta_x)}\right) + \sum_{y \in [k] \backslash \{x\}} \Qbf_{Q_0}(y|\delta_x) f(0)\\
    & = \Qbf_{Q_0}(x|\delta_x) f\left(\frac{1}{\Qbf_{Q_0}(x|\delta_x)}\right) + (1-\Qbf_{Q_0}(x|\delta_x)) f(0). \label{eq:WorstfDivFiniteLB}
\end{align}
We can observe that the last term can be written in the form of the optimal worst-case $f$-divergence \eqref{eq:optWorstfDivFormulaOnly}
with $r_1=0$ and $r_2=1/\Qbf_{Q_0}(x|\delta_x)$. In other words, let
\begin{align}
    \mathfrak{R}(r_1,r_2) = \frac{1-r_1}{r_2-r_1} f(r_2) + \frac{r_2 - 1}{r_2 - r_1} f(r_1).
\end{align}
Then we have $R_f(\Qbf_{Q_0}) \geq \mathfrak{R}(0, 1/\Qbf_{Q_0}(x|\delta_x))$.

As noted in Appendix \ref{supp:mechBehavior}, $\mathfrak{R}(r_1,r_2)$ is increasing in $r_2$ for $0 \leq r_1 < 1 < r_2$. Since $\Qbf_{Q_0}(x|\delta_x) \leq B(Q_0(x))$, we have
\begin{align}
    R_f(\Qbf_{Q_0}) \geq \mathfrak{R}(0,{1/B(Q_0(x))}).
\end{align}
Since this should hold for all $x \in [k]$, we have
\begin{align}
    R_f(\Qbf_{Q_0}) \geq \max_{x \in [k]} \mathfrak{R}(0,{1/B(Q_0(x))}).
\end{align}
Since $\min_{x \in [k]} Q_0(x) \leq 1/k$ for all $Q_0 \in \Pcal([k])$, and $B(t)$ is increasing in $t$, we have
\begin{align}
    R_f(\Qbf_{Q_0}) \geq \mathfrak{R}(0,1/B(1/k)).
\end{align}
Now, let $\mu_k$ be the uniform distribution over $[k]$. We will show that 
\begin{align}
     R_f(\Qbf_{\mu_k}) = \mathfrak{R}(0,1/B(1/k)),
\end{align}
which suffices to prove that $Q_0=\mu_k$ minimizes $R_f(\Qbf_{Q_0})$.

We observe that $\Mcal_{\epsilon, \mu_k}$ is a convex set in $\Pcal([k])$. Since $\fDiv{P}{Q}$ is jointly convex in $(P,Q)$, we obtain that $\fDiv{P}{\Qbf_{\mu_k}(P)} = \min_{Q \in \Mcal_{\epsilon, \mu_k}} \fDiv{P}{Q}$ is convex in $P$ by \cite[Section 3.2.5]{Boyd04-convOptBook}. Hence, the maximum of $\fDiv{P}{\Qbf_{\mu_k}(P)}$ occurs when $P$ is one of the extreme points of $\Pcal([k])$, that is, the point masses $\delta_x$. By the same arguments as in \eqref{eq:WorstfDivFiniteLBInit}-\eqref{eq:WorstfDivFiniteLB}, we have $\fDiv{\delta_x}{Q} = \mathfrak{R}(0, 1/Q(x))$, and hence
\begin{align}
    R_f(\Qbf_{\mu_k}) &= \sup_{P \in \Pcal([k])} \fDiv{P}{\Qbf_{\mu_k}(P)}\\
    &= \max_{x \in [k]}\fDiv{\delta_x}{\Qbf_{\mu_k}(\delta_x)} \\
    &= \max_{x \in [k]} \inf_{Q \in \Mcal_{\epsilon, \mu_k}} \fDiv{\delta_x}{Q} \\
    &= \max_{x \in [k]} \inf_{Q \in \Mcal_{\epsilon, \mu_k}} \mathfrak{R}(0, 1/Q(x)) \\
    &= \mathfrak{R}\left(0, \frac{1}{\min_{x \in [k]} \sup_{Q \in \Mcal_{\epsilon, \mu_k}} Q(x)}\right).
\end{align}
Also, by the same arguments as in \eqref{eq:QBoundPrevInit}-\eqref{eq:QBoundPrevFinal}, we have
\begin{align}
    Q(x) \leq B(1/k), \quad \forall Q \in \Mcal_{\epsilon, \mu_k}, x \in [k].
\end{align}
Now, we will show that $\sup_{Q \in \Mcal_{\epsilon, \mu_k}} Q(x) = B(1/k)$ for any $x \in [k]$. To show this, we will prove that there is a distribution $Q \in \Pcal([k])$ such that $Q(x)=B(1/k)$ and $Q(y) = \frac{1-B(1/k)}{k-1}$ for all $y \in [k] \backslash \{x\}$, and this $Q$ is contained in $\Mcal_{\epsilon, \mu_k}$. It suffices to show the followings:
\begin{align}
    \frac{1}{k}e^{-\epsilon/2} \leq B(1/k) \leq \min\left\{1, \frac{1}{k}e^{\epsilon/2}\right\},\\
    \frac{1}{k}e^{-\epsilon/2} \leq \frac{1-B(1/k)}{k-1} \leq \min\left\{1, \frac{1}{k}e^{\epsilon/2}\right\}.
\end{align}

We note that whenever $0 \leq t \leq 1$, we have
\begin{itemize}
    \item $B(t)=\min\{e^{\epsilon/2}t, e^{-\epsilon/2}t + (1-e^{-\epsilon/2})\} \geq t$, and
    \item $B(t) \leq e^{-\epsilon/2}t + (1-e^{-\epsilon/2}) = 1-(1-t)e^{-\epsilon/2} \leq 1$.
\end{itemize}
 From these, we can easily observe that $B(1/k) \leq \min\left\{1, \frac{1}{k}e^{\epsilon/2}\right\}$, and
\begin{align}
    \frac{1}{k}e^{-\epsilon/2} \leq \frac{1}{k} \leq B(1/k).
\end{align}
Also, 
\begin{gather}
    \frac{1-B(1/k)}{k-1} \leq \frac{1-(1/k)}{k-1} = \frac{1}{k} \leq \min\left\{1, \frac{1}{k}e^{\epsilon/2}\right\},\\
    \frac{1-B(1/k)}{k-1} \geq \frac{1-[e^{-\epsilon/2}(1/k) + (1-e^{-\epsilon/2})]}{k-1}=\frac{1}{k}e^{-\epsilon/2}.
\end{gather}
This shows that $\sup_{Q \in \Mcal_{\epsilon, \mu_k}} Q(x) = B(1/k)$ for any $x \in [k]$. Thus,
\begin{align}
    \min_{x \in [k]} \sup_{Q \in \Mcal_{\epsilon, \mu_k}} Q(x) &= B(1/k),\\
    R_f(\Qbf_{\mu_k}) &= \mathfrak{R}\left(0, \frac{1}{B(1/k)}\right).
\end{align}
This ends the proof that $Q_0=\mu_k$ minimizes $R_f(\Qbf_{Q_0})$, and $R_f(\Qbf_{\mu_k}) = \mathfrak{R}(0,1/B(1/k))$.

\subsection{Explanation for the experiment in 1D Gaussian mixture}\label{supp:numResultSetupsGaussMix}
The precise description of the set of possible client distributions in our experimental setup is as follows:
\begin{align}
    \PtildCal = \left\{P: p(x) = \frac{\sum_{i=1}^{k} \lambda_i \mathcal{N}_{\mu_i,1}[x]}{\int_{-4}^{4} \sum_{i=1}^{k} \lambda_i \mathcal{N}_{\mu_i,1}[x] dx} \mathbbm{1}_{[-4,4]}(x), k \in [K], \lambda_i \geq 0, \sum_{i=1}^{k}\lambda_i = 1, |\mu_i| \leq 1 \right\}.\label{eq:GaussMixExpSetup}
\end{align}

Each of $P_j \in \PtildCal$ is generated by choosing $k, \lambda_i$, and $\mu_i$ in \eqref{eq:GaussMixExpSetup} as follows: (i) First, choose the number of Gaussian distributions $k$ by first sample $\tilde{k}$ from the Poisson distribution with mean $k_0$ (which is chosen beforehand), and let $k = \min(\tilde{k}+1, K)$, and (ii) after that, sample each of $\mu_1,\cdots,\mu_k$ independently from the uniform distribution on $[-1,1]$, and sample $(\lambda_1,\cdots,\lambda_k)$ from the uniform distribution on $\Pcal([k])$.

The implementation of our proposed mechanism is as follows. We can observe that
\begin{align}
    \int_{-4}^{4} \sum_{i=1}^{k} \lambda_i \mathcal{N}_{\mu_i,1}[x] dx
    \geq  \inf_{\mu \in [-1,1]}  \int_{-4}^{4} \mathcal{N}_{\mu,1}[x] dx
    =  \int_{-4}^{4} \mathcal{N}_{1, 1}[x] dx = \Phi(3)-\Phi(-5),
\end{align}
where $\Phi$ is the cdf of the 1D standard Gaussian distribution, and for each $x \in [-4,4]$, we have
\begin{align}
    \sum_{i=1}^{k} \lambda_i \mathcal{N}_{\mu_i,1}[x] dx \leq \sup_{\mu \in [-1,1]} \mathcal{N}_{\mu,1}[x] = \frac{1}{\sqrt{2\pi}} \exp\left(-[\max(|x|-1, 0)]^2/2\right).
\end{align}
Hence, we have $\PtildCal \subset \PtildCal_{0,1,\tilde{h}} = \PtildCal_{0, c_2, h}$, where 
\begin{align}
    \tilde{h}(x) = \frac{\exp\left(-[\max(|x|-1, 0)]^2/2\right)}{\sqrt{2\pi} (\Phi(3)-\Phi(-5))} \mathbbm{1}_{[-4,4]}(x)
\end{align}
and $c_2 = \int \tilde{h}(x) dx$, $h(x)=\tilde{h}(x)/c_2$. 
When implementing our proposed mechanism, we use the bisection method to find a constant $r_P$.
We predetermine the error tolerances $\delta_1, \delta_2 \geq 0$, $\delta_1 < 1$, and we terminate the bisection method to find $r_P$ if the integration of \eqref{eq:contMech} lies in the interval $[1-\delta_1,1+\delta_2]$.
As mentioned in Section \ref{subsec:effApproxOfConstFactor}, to consider the error tolerances, we actually implement $\Qbf^*_{0,c_2,h,\epsilon'}(P)$, with $\epsilon'=\epsilon - \log\frac{1+\delta_2}{1-\delta_1}$.

In the experiment in the paper, we use $k_0=2$ and $\delta_1=\delta_2=10^{-5}$.

For the baseline, we use the same hyperparameter setups as in \cite[Section 5, Paragraph ``Architectures"]{Husain20-LDPSampling}, except a slight modification in a reference distribution $Q_0$. In \cite{Husain20-LDPSampling}, they set the standard Gaussian as the reference distribution, $Q_0=\mathcal{N}(0, 1)$. To consider the truncated domain $[-4,4]$, we set $Q_0$ as the truncation of the standard Gaussian, that is, $Q_0$ has a pdf
\begin{align}
    q_0(x) = \frac{\mathcal{N}_{0,1}[x]}{\int_{-4}^{4} \mathcal{N}_{0,1}[x] dx} \mathbbm{1}_{[-4,4]}(x).
\end{align}

The baseline has many other hyperparameters not specified in \cite{Husain20-LDPSampling}, such as the proposal distribution used for the Metropolis-Hasting algorithm \cite{Metropolis53-EqStateCalcFastCompMachine, Robert16-MHAlgo}, initial model parameters for the weak learner, etc. 
We noticed that the code for \cite{Husain20-LDPSampling} is available at \url{https://github.com/karokaram/PrivatedBoostedDensities/tree/master}
and hence we made every effort to faithfully reproduce the baseline with  exactly the same hyperparameters, including those not mentioned in the paper \cite{Husain20-LDPSampling}. However, subtle variations may arise due to differences in the programming languages employed; we used Python, while \cite{Husain20-LDPSampling} used Julia.

\subsection{Explanation for Figure \ref{fig:ringGauss}}
For $k \in \Nbb$ and $\sigma>0$, the Gaussian ring distribution with $k$ modes and a component-wise variance $\sigma^2$ is the mixture of $k$ Gaussian distributions in $\Rbb^2$ with equal weights, where each Gaussian distribution has the covariance $\sigma^2 I_2$ and equally spaced mean in the unit circle, and one of the Gaussian distribution has mean $(1,0)$. That is, it has a density $\frac{1}{k} \sum_{i=1}^{k}\mathcal{N}_{\mu_i,\sigma^2 I_2}[x]$, with $\mu_i = (\cos \frac{2\pi i}{k}, \sin \frac{2\pi i}{k})$. In Figure \ref{fig:ringGauss}, the left image is the pdf of the Gaussian ring distribution with $k=3$ and $\sigma^2=0.5$. It is used as a client's original distribution.

For our proposed mechanism, similar to Section \ref{subsec:contSpace}, we use the setup that $\PtildCal$ consists of Gaussian mixtures, where each Gaussian has mean within a unit ball centered at the origin and has covariance $0.5 I_2$. that is, $\PtildCal = \{\sum_{i=1}^{k} \lambda_i \mathcal{N}(x_i, \sigma^2 I_{2}): k \in \Nbb, \lambda_i \geq 0, \sum_{i=1}^{k}\lambda_i = 1, \Vert x_i \Vert \leq 1\}$, with $\sigma^2=0.5$. We can also observe that $\PtildCal \subset \PtildCal_{0,1,\tilde{h}}$ for $\tilde{h}(x)=\frac{1}{2\pi \sigma^2} \exp\left(-\frac{[\max(0, \Vert x \Vert - 1)]^2}{2\sigma}\right)$.

Same as Appendix \ref{supp:numResultSetupsGaussMix}, we have $\PtildCal_{0,1,\tilde{h}}=\PtildCal_{0, c_2, h}$, where $c_2 = \int \tilde{h}(x) dx$, $h(x)=\tilde{h}(x)/c_2$. Hence, we use $\Qbf^*_{0,c_2,h,\epsilon}$. The implementation of $\Qbf^*_{c_1,c_2,h,\epsilon}(P)$ is the same as the description in Appendix \ref{supp:numResultSetupsGaussMix} with $\delta=10^{-5}$. For the baseline, we also use MBDE with the same hyperparameter setup as \cite[Section 5, Paragraph ``Architectures"]{Husain20-LDPSampling}, with the (untruncated) 2D standard Gaussian as a reference distribution, $Q_0=\mathcal{N}(0, I_2)$.

\section{Additional Numerical Results for Finite Space} \label{supp:addExp}
In this appendix, we present more numerical results for the finite space case explained in Section \ref{subsec:numResultFinSpace} for other $k$. We present the result for $k=5,20$, and $100$ in Figures  \ref{fig:expFinSpaceK5}-\ref{fig:expFinSpaceK100}. As shown by the figures, the proposed mechanism always has the smaller worst-case $f$-divergence compared to the baseline, as the optimality is proved for the proposed mechanism. However, the performance gap between two mechanisms decreases as $\epsilon$ becomes smaller and $k$ becomes larger.

\begin{figure}[htbp]
    \centering
    \includegraphics[width=0.8\textwidth]{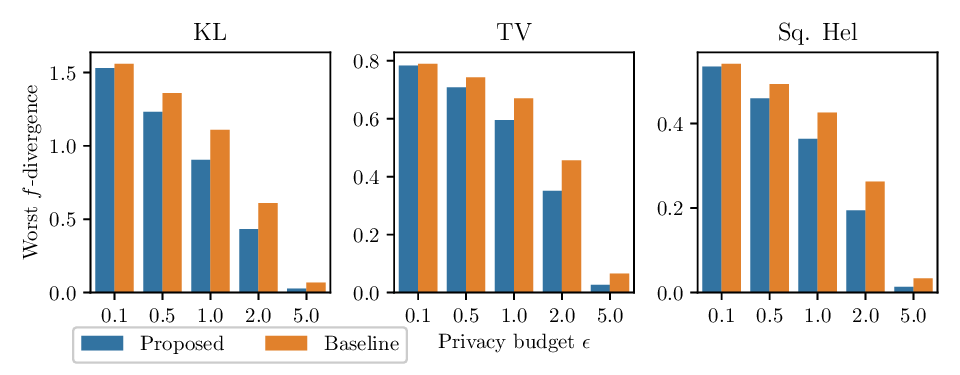}
    \caption{Theoretical worst-case $f$-divergences of proposed and previously proposed baseline mechanisms (with uniform $Q_0$) over finite space ($k=5$)}
    \label{fig:expFinSpaceK5}
\end{figure}

\begin{figure}[htbp]
    \centering
    \includegraphics[width=0.8\textwidth]{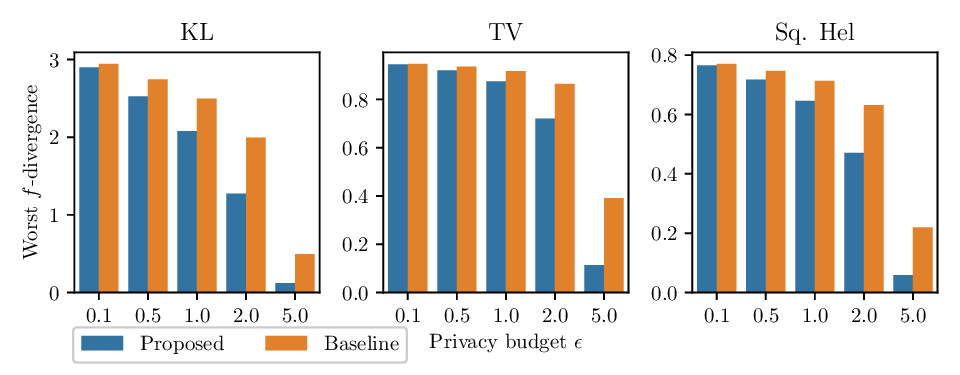}
    \caption{Theoretical worst-case $f$-divergences of proposed and previously proposed baseline mechanisms (with uniform $Q_0$) over finite space ($k=20$)}
    \label{fig:expFinSpaceK20}
\end{figure}

\begin{figure}[htbp]
    \centering
    \includegraphics[width=0.8\textwidth]{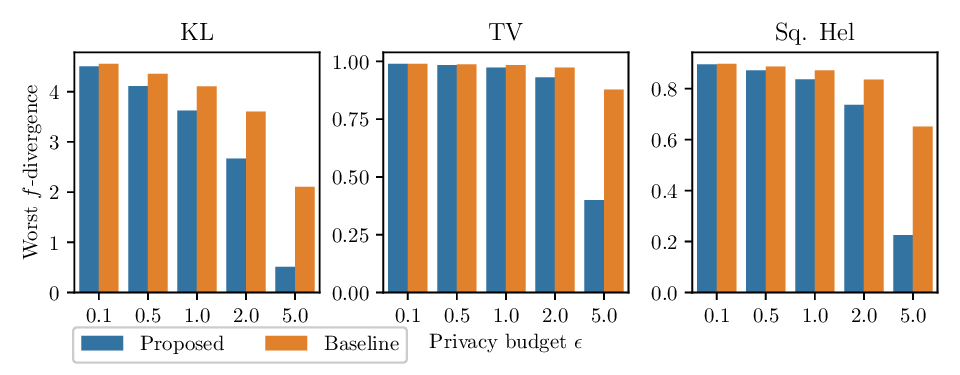}
    \caption{Theoretical worst-case $f$-divergences of proposed and previously proposed baseline mechanisms (with uniform $Q_0$) over finite space ($k=100$)}
    \label{fig:expFinSpaceK100}
\end{figure}

\section{Instructions for Reproducing Results} \label{supp:reprodInst}
In this appendix, we provide instructions for reproducing the experiments and figures in the paper. 
For a detailed document about the code, refer to the provided file  \verb|README.md|.
For the tasks consuming a large time, we also specify the running times of such tasks. We do not specify the running times of the short tasks consuming less than 5 seconds. All experiments are performed on our simulation PC with the following specifications:
\begin{itemize}
    \item OS: Ubuntu 22.04.1
    \item CPU: Intel(R) Core(TM) i9-9900X
    \item Memory: 64GB
\end{itemize}

There are a few remarks:
\begin{itemize}
    \item Although we expect that the codes can be run and reproduce our results on sufficiently recent versions of Python libraries, we provide the information about the anaconda environment used in the experiment in \verb|environment.yaml| for the completeness.
    \item The provided codes contain some lines to make the figures use \TeX{} fonts. Running such lines requires the \LaTeX{} to be installed in the experimental environment. We can remove the following lines to disable using \TeX{}:
    \begin{verbatim}
        matplotlib.use("pgf")

        "pgf.texsystem": "pdflatex",
        'font.family': 'serif',
        'text.usetex': True,
        'pgf.rcfonts': False,
        'font.serif' : 'Computer Modern Roman',
    \end{verbatim}
\end{itemize}

\subsection{Instruction for producing Figure \ref{fig:ringGauss}}
Figure \ref{fig:ringGauss} can be obtained by running the code \verb|plot_GaussRing.py|, without any program arguments. We measure the running time for initializing the mechanism and and calculating the sampling distribution for the baseline (MBDE \cite{Husain20-LDPSampling}) and our proposed mechanism. The measured running times in our environment are as follows: (unit: second)
\begin{itemize}
    \item Initializing the mechanism
    \begin{itemize}
        \item Baseline: 0.80
        \item Proposed: 1.22
    \end{itemize}
    \item Calculating the sampling distribution
    \begin{itemize}
        \item Baseline: 35.19
        \item Proposed: 664.18
    \end{itemize}
\end{itemize}

\subsection{Instruction for producing Figure \ref{fig:discMechVisual}}
Figure \ref{fig:discMechVisual} can be obtained by running the code \verb|visualize_finiteSpace.py|, without any program arguments.

\subsection{Instruction for producing results for finite space}
The results about the theoretical worst-case $f$-divergence for finite space, Figures \ref{fig:expFinSpace},\ref{fig:expFinSpaceK5},\ref{fig:expFinSpaceK20},\ref{fig:expFinSpaceK100}, can be obtained by running the code \verb|plot_finite.py| with a program argument \verb|--k| to specify the value of $k$. For example, in the command line, the aforementioned four figures can be generated by the following commands, respectively:
\begin{verbatim}
python plot_finite.py --k 10
python plot_finite.py --k 5
python plot_finite.py --k 20
python plot_finite.py --k 100
\end{verbatim}

\subsection{Instruction for producing results for 1D Gaussian mixture}
The experiment of 1D Gaussian mixture in Section \ref{subsec:numResult1DGaussMix} consists of the following two codes:
\begin{enumerate}
    \item \verb|exp_1DGaussMix.py|

    This code performs an experiment on a single $\epsilon$. We can provide two program arguments \verb|--eps| and \verb|--seed| to specify the values of $\epsilon$ and the random seed, respectively.

    \item \verb|plot_1DGaussMix.py|

    This code generate the plot like Figure \ref{fig:expGaussMixture1D} from the results of \verb|exp_1DGaussMix.py|
\end{enumerate}

The results in the paper, Figure \ref{fig:expGaussMixture1D}, can be obtained as follows:
\begin{enumerate}
    \item First, run the following commands:

    \begin{verbatim}
        python exp_1DGaussMix.py --eps 0.1 --seed 1
        python exp_1DGaussMix.py --eps 0.5 --seed 2
        python exp_1DGaussMix.py --eps 1.0 --seed 3
        python exp_1DGaussMix.py --eps 2.0 --seed 4
        python exp_1DGaussMix.py --eps 5.0 --seed 5
    \end{verbatim}

    These can be run in any order or in parallel.
    Check that all of the five result files \verb|data_1DGaussMix_eps{eps}.npy| corresponding to five values of $\epsilon$ are generated. In our environment, running all of the five commands in parallel consumes 3h 13m 35s.

    \item Then, run the code \verb|plot_1DGaussMix.py| without any program arguments.
\end{enumerate}

\section{Limitations}\label{supp:limitations}
Our main contribution lies in proposing a minimax-optimal mechanism, and we present several experimental results based on synthetic data to support the superiority of our mechanism. However, since we have not conducted experiments based on real datasets, the analysis of performance in real-world scenarios is insufficient. 
Also, our PUT formulation in the minimax sense provides optimal utility in the worst case, which may result in reduced \emph{average} utility when prior information is given.
Finally, our implementation of the proposed mechanism requires a large amount of running time due to numerical integration, which makes experiments in multidimensional spaces challenging. 

\section{Broader Impacts}\label{supp:broaderImpacts}
Our proposed mechanism can be utilized for privacy protection in the field of generative models, which has recently received significant attention. A major deterrent to the adoption of privacy protection algorithms in real-world scenarios is the potential performance degradation. Our PUT-optimal mechanism, that minimizes the loss of utility given the privacy budget, can alleviate such concerns. 

We should note that an LDP mechanism provides a certain level of privacy protection but cannot guarantee complete privacy protection without completely sacrificing utility. Additionally, clients typically provide multiple data points through various channels, and when these data are combined, it can lead to greater privacy leakage \cite{Kairouz15-CompThmDP, Rogers16-PrivOdometFilter}.

\end{document}